%% file: PD-ML.tex
\documentclass{article}

\PassOptionsToPackage{numbers}{natbib}
\usepackage[numbers]{natbib}

\usepackage[utf8]{inputenc} 
\usepackage[T1]{fontenc}    
\usepackage{hyperref}       
\usepackage{url}            
\usepackage{booktabs}       
\usepackage{amsfonts}       
\usepackage{nicefrac}       
\usepackage{microtype}      

\title{PD-ML-Lite: Private Distributed Machine Learning from Lightweight Cryptography}

\newcommand{\R}{\mathbb R}

\newcommand{\remove}[1]{}

\author{Maksim Tsikhanovich\thanks{Bloomberg LP, {\tt maksimtsikhanovich@gmail.com}}\ $^\S$ \and Malik Magdon-Ismail\thanks{Computer Science Department, RPI, {\tt magdon@gmail.com}} \and Muhammad Ishaq\thanks{School of Informatics, University of Edinburgh, {\tt s1798792@sms.ed.ac.uk}.}
  $^\P$ \and  Vassilis Zikas\thanks{School of Informatics, University of Edinburgh, {\tt vzikas@inf.ed.ac.uk}.}\ \thanks{Work done in part while the author was at RPI.}}

\usepackage{amsmath}
\usepackage{amsthm}
\usepackage{fullpage}
\usepackage{wrapfig}
\usepackage{setspace}
\usepackage{url}            
\usepackage{booktabs}       
\usepackage{amsfonts}       
\usepackage{nicefrac}       
\usepackage{microtype}      
\usepackage{ulem}
\usepackage{multicol}
\usepackage{enumerate}
\usepackage{multirow} 
\usepackage{amsmath, amssymb,amsthm}
\usepackage{mathtools}
\usepackage{latexsym}
\usepackage{moreverb} 
\usepackage{graphicx} 
\usepackage{array} 
\usepackage{xcolor,colortbl}
\usepackage{booktabs}
\usepackage{xspace}
\usepackage{relsize}
\usepackage{paralist}

\usepackage{tikz}
\usetikzlibrary{arrows,arrows.meta}
\usetikzlibrary{decorations.markings}
\usetikzlibrary{decorations.pathmorphing}
\usetikzlibrary{positioning,shapes,backgrounds}
\usetikzlibrary{calc,intersections,patterns,fit,automata}
\usepackage{tikzsymbols}

\usepackage{algorithm}
\usepackage{algorithmicx}
\usepackage[noend]{algpseudocode} 
    
\newenvironment{tightcenter}
 {\parskip=0pt\par\nopagebreak\centering}
 {\par\noindent\ignorespacesafterend}

\def\E{\cl{E}}
\def\SMSP{\ssum}
\def\argmin{\mathop{\arg\,\min}}
\def\Prob{{\mathbb{P}}}

\newtheorem{theorem}{Theorem}
\newtheorem{claim}{Claim}

\newcounter{ctr}

\newcounter{ectr}
\input{macros.tex}

\begin{document}

\maketitle

\vspace{-2ex}\begin{abstract}
  \input{abstract.tex}
\end{abstract}

\input{introduction.tex}
\input{algorithm.tex}

\input{algorithm_svd.tex}

\input{privacy_analysis.tex}

\input{experimental_results.tex}
\input{experimental_results_svd.tex}

\input{conclusion.tex}

\newpage
\appendix

\input{appendix.tex}

\input{supp_svd.tex}
%

\clearpage
\bibliographystyle{abbrvnat}
{
\bibliography{biblio,biblio_svd}
}

\end{document}

%% file: macros.tex
\newcommand{\pdsvd}{\ensuremath{\text{\sc PD-SVD}}\xspace}
\newcommand{\ssum}{\ensuremath{\text{\sc SecSum}}\xspace}
\newcommand{\nssum}{\ensuremath{\text{\sc NormedSecSum}}\xspace}

\newcommand{\termdef}[1]{\textit{#1}\xspace}

\renewcommand{\algorithmicrequire}{\textbf{Input:}}
\renewcommand{\algorithmicensure}{\textbf{Output:}}

\def\math#1{$#1$}

\def\mand#1{$$#1$$}
\def\mandc#1{\mand{\abovedisplayskip=3pt plus 1pt minus 1pt%
\abovedisplayshortskip=0pt plus 1pt minus 1pt%
\belowdisplayskip=3pt plus 1pt minus 1pt%
\belowdisplayshortskip=0pt plus 1pt minus 1pt%
#1}}

\def\frac#1#2{{#1\over #2}}

\def\mld#1{\begin{equation}
#1
\end{equation}}
\def\mldc#1{\mld{\abovedisplayskip=3pt plus 1pt minus 1pt%
\abovedisplayshortskip=0pt plus 1pt minus 1pt%
\belowdisplayskip=3pt plus 1pt minus 1pt%
\belowdisplayshortskip=0pt plus 1pt minus 1pt%
#1}}

\DeclareSymbolFont{AMSb}{U}{msb}{m}{n}
\DeclareMathSymbol{\N}{\mathbin}{AMSb}{"4E}
\DeclareMathSymbol{\Q}{\mathbin}{AMSb}{"51}
\DeclareMathSymbol{\I}{\mathbin}{AMSb}{"49}
\DeclareMathSymbol{\C}{\mathbin}{AMSb}{"43}

\def\qed{\hfill\rule{2mm}{2mm}}

\def\cl#1{{\cal #1}}

\def\mbf#1{{\mbox{\boldmath{$#1$}}}}

\def\norm#1{{\|#1\|}}

\def\r#1{{(\ref{#1})}}

\theoremstyle{definition}
\newtheorem{remark}{\bf Remark}

\newcounter{exercisenum}

\def\dotfil{\leaders\hbox to 1.5mm{.}\hfill}
\newcounter{rmnum}
\def\RN#1{\setcounter{rmnum}{#1}\uppercase\expandafter{\romannumeral\value{rmnum}}}
\def\rn#1{\setcounter{rmnum}{#1}\expandafter{\romannumeral\value{rmnum}}}

\newlength{\saveparindent}
\newlength{\saveparskip}
\newenvironment{tiret}{%
\begin{list}{\hspace{1pt}\rule[0.5ex]{6pt}{1pt}\hfill}{\labelwidth=15pt%
\labelsep=3pt \leftmargin=18pt \topsep=1pt%
\setlength{\listparindent}{\saveparindent}%
\setlength{\parsep}{\saveparskip}%
\setlength{\itemsep}{1pt}}}{\end{list}}

\newcommand{\m}{^{(m)}}
\newcommand{\para}[1]{\par\textbf{#1}\hspace{1em}}
\newcommand{\ksdp}{KSDP\xspace}

%% file: abstract.tex
Privacy is a major issue in learning from distributed data.
Recently
the cryptographic literature has provided several tools for this task.
However, these tools either reduce the quality/accuracy of the learning algorithm---e.g., by adding noise---or they incur a high performance penalty and/or involve trusting external authorities. 

We propose a methodology for {\sl private distributed machine learning from light-weight cryptography} (in short, PD-ML-Lite). We apply our methodology to two major ML algorithms, namely non-negative matrix factorization (NMF) and singular value decomposition (SVD). 
 Our resulting protocols  are communication optimal, achieve the same accuracy as their non-private counterparts, and satisfy a notion of privacy---which we define---that is both intuitive and measurable. Our approach is to
 use lightweight cryptographic protocols (secure sum and normalized secure sum)
 to build learning algorithms rather than wrap complex learning algorithms
 in a heavy-cost MPC framework.
 
We showcase our algorithms' utility and privacy on several applications:
 for NMF we consider topic modeling and recommender systems, and for SVD, principal component regression, and low rank approximation.

%% file: introduction.tex
\section{Introduction}

More data is better for all forms of learning.
In 
many domains the data is distributed among several parties, 
so the best learning outcome requires data sharing.
However, sharing raw data between organizations can be uneconomic
and/or subject to policies or even legislation.
For example, consider the following (distributed) ML application scenarios:
\begin{tiret}
\item[{\it Private Distributed Topic Modeling.}]
Government agencies with protected data (FBI, CIA, NSA, \ldots)
wish to build a distributed peer-to-peer information retrieval system, and doing
so requires a topic model over all their data \cite{tang2003peer,yi2009comparative}. 
\item[{\it Private Distributed Recommender Systems.}]
Businesses with proprietary consumer data would like to build 
recommender systems which can leverage data
across all the businesses without compromising the privacy of any party's
data \cite{5477138}.
\end{tiret}
In such applications, a key constraint is that each organization already has a system that serves their needs to some extent.
Thus an organization's motivation for participating in distributed learning is in {\sl improving} the quality of their local system while ensuring their data remains {\sl private.}

In this work, we focus on two fundamental tools in machine learning for computing
compact, useful representations of a data matrix: non-negative matrix factorization (NMF) and singular value decomposition (SVD).
NMF is NP-hard~\cite{Vavasis2009-NMF-Hardness}, while
SVD can be solved in cubic time~\cite{golub2012matrix}.
For both of these tasks, 
there are families of algorithms and heuristics with
acceptable performance in practice, e.g., for NMF~\cite{lin2007projected,cichocki2007hierarchical,honmf} and for SVD~\cite{golub2012matrix}.
Many of these NMF heuristics can be
performed efficiently even when the data is distributed~\cite{gemulla2011large,du2014maxios,berry2003parallel}, and there is significant interest on
distributed algorithms for SVD (e.g. ~\cite{IwenOng2016}).
The typical focus of such distributed algorithms is to minimize the information
that has to be communicated to obtain the same outcome as a centralized algorithm,
so these algorithms are not privacy preserving --- a critical challenge which we address here. Stated simply, the main question we attack is as follows 
\begin{center}
\parbox{0.775\linewidth}{\it How can the parties
  collectively learn from  data distributed amongst themselves,
  without needing a trusted third-party, while ensuring that: (1)  little information is leaked from any one party
        to another, and (2) the learning outcome is comparable, ideally identical, to that which could be obtained if all the data were centralized?}
\end{center}

\subsection{\bf Privacy in (Distributed) Learning.}
Data privacy is an emerging topic in machine-learning.
Naive solutions like data anonymization are 
insufficient~\cite{brickell2008cost,malin2013biomedical}. Recently, the cryptographic community has investigated the application of cryptographic techniques to enhance the privacy of learning algorithms. The two main methods in this realm  are: {\sl differential privacy (DP)} and {\sl secure multi-party computation (MPC).}
These
methods differ not only on how they are implemented,
but most importantly on their use cases,
their security/privacy guarantees, and their computation overhead. 

In a nutshell, using DP one can respond to a query on a database while
preserving privacy of any individual record against an {\sl external}
observer that sees the output of the query (no matter what side-information he  has on the data.) This is achieved by  a trusted curator who has access to the data
adding noise to the output, which "flattens" any individual record's  impact on the output, thereby hiding that record. On the other hand, MPC  allows $n$ mutually distrustful parties, each holding a private dataset, to compute {and output} a function on their joint data without leaking {\sl to each other} information other than the output. This privacy guarantee, which
is also what we aim for in this work, is orthogonal to DP: MPC computes
the exact output and does not protect the privacy of individual records
from what might be inferrable from the output. Thus, the decision to use DP, MPC, or something else must consider the learning and
privacy requirements of the application.
In Appendix~\ref{appendix:MPCvsDP} we include an informal table comparing
MPC and DP from the scope of ML. 
We now argue why each of these solutions is inadequate
for our purposes.

\paragraph{\bf Unsuitability of DP.} A fundamental constraint is that the
{\it learning outcome of the private distributed algorithm must match
the centralized\footnote{Centralized refers to the optimal (non-private)
outcome
where all data is aggregated for
learning.} outcome},
because the \emph{only} reason for the distributed parties to
collaborate is if they can improve on the local learning outcome in terms
of prediction performance. 
This \emph{renders DP unsuitable for our goals,}  as the addition of noise inevitably deteriorates the learning outcome~\cite{rajkumar2012differentially,blum2005practical,rajkumar2012differentially,wang2015privacy, FastDPMF, hua2015differentially}.   Note that for our applications the strict accuracy restriction is particularly important. For example, if adding privacy renders the  distributed recommender system less accurate than a system one party could construct from its own data, then this party gained nothing and lost some privacy by participating in distributed learning. For completeness we include in Figure~\ref{fig:learning_outcome}(c)
and Figure~\ref{fig:svd_learning_uplift} (see Section~~\ref{sec:experimental_results})
experimental results 
confirming general observations in the literature
that adding noise to accommodate even a moderate level of differential
privacy results in excessive
deterioration in accuracy for the distributed NMF and SVD algorithms
(also see supplemental material for details).

\paragraph{\bf Insufficiency of general MPC.}
In theory, MPC can solve our problem exactly: One can run an arbitrary optimization/inference algorithm on distributed data by executing some type of
`MPC byte code' on a `distributed virtual machine.' 
Such an algorithm-agnostic implementation typically \emph{incurs a high communication and computation overhead,} e.g., ~\cite{nikolaenko2013privacy}. 
More recently, 
optimized MPC protocols tailored to private machine learning on distributed
data~\cite{Kim2016,mohasselsecureml,DBLP:conf/ccs/MohasselR18,DBLP:conf/ccs/MazloomG18} were developed. However, these rely on the assumption of  
partially trusted third  parties that take on
the burden of the computation and privacy protection. This makes privacy more tractable, but it is arguably a strong assumption for practical applications, a  compromise which we do not make here. 

\subsection{Our Contributions.}
In this work, we propose  a paradigm shift in combining cryptography with machine learning. At a high level, instead of relying on cryptography developing (new) MPC for machine learning, we identify small distributed operations that can be performed securely and very efficiently using existing lightweight cryptographic tools---in particular, elementary MPC operations such as parallel additions and multiplications/divisions---and develop new learning algorithms that only communicate by means of such tools. We note that although these tools are cryptographically secure,  i.e., they leak nothing about their inputs, they do announce their output, which is a (typically obfuscated) function of the inputs. To quantify the privacy loss incurred by announcing such functions, we adapt ideas from the noiseless privacy-preserving data-release literature. We note in passing that the idea of estimating privacy loss from a leaky MPC, by means of a privacy-preservation mechanisms, in particular DP, was recently also used in~\cite{CCS:MazGor18}, for the model where the computation is outsourced to two parties that are semi-trusted (one of them must be honest). We do not make any such requirement here. 

We apply our methodology to two classical problems in machine learning which are highly relevant for a spectrum of applications, namely,  Non-negative Matrix Factorization (NMF) and  singular value decomposition (SVD). We further demonstrate the performance and privacy of our algorithms for each of these problems to typical applications.

\subsubsection{Overview of Private Distributed NMF}
We give an algorithm for
private, distributed Non-negative Matrix Factorization (PD-NMF). In a nutshell, our algorithm allows $M$ parties, where each party $i$ holds as input a database
$X^{(i)}$ (a non-negative matrix), to distributedly compute a solution to NMF
on the union of their inputs \math{\cup_{i}X^{(i)}},
without exchanging their input-databases or revealing considerable information about any individual record in their databases other than the output of the NMF.
Importantly, we guarantee that
the output of the NMF should be the same as computing a centralized NMF,
where an imaginary trusted third party collets all data, runs the NMF
algorithm and then distributes the result to each party.
\footnote{Recall that this requirement renders DP mechanisms
  unacceptable,  as adding noise would incur an error on the output.}
We refer to such an idealized protocol that uses this imaginary
trusted third party for solving the problem as {\sl centralized NMF}.
The formulation of the PD-NMF problem is given in Figure~\ref{fig:prob-statement}.  

\begin{figure}[ht!]
\centerline{\fboxsep5pt\fbox{\parbox{0.8\linewidth}{
\textbf{PD-NMF: Private Distributed NMF.}
{
Each of $M$ mutually distrustful parties have a non-negative
\math{n_m\times d} matrix $X^{(m)}$ ($n_m$ rows over $d$ features),
for \math{m=1,\ldots,M}.
The parties agree on the  $d$ features
  and their order. 
Given a rank $k$,
each party must compute the same non-negative row-basis matrix
$T\in\R^{k\times d}$
such that:
\\[1pt]
{\tabcolsep2pt
\begin{tabular}{p{0.02\linewidth}p{0.95\linewidth}}
\math{\bullet}&
The sum of Frobenius reconstruction errors for each matrix
\math{X^{(m)}} onto the basis \math{T} plus an optional regularization term
is minimized:
\mldc{\mathop{\text{min}}\limits_{T}\ \textstyle\frac{1}{2}\sum\limits_{m=1}^M ||X^{(m)}-W^{(m)}T||_F^2 + \text{reg}(W,T),
\label{eq:1}}
where $W^{(m)}\in\R^{n_m\times k}$ is an optimal non-negative least-squares
fit of \math{X^{(m)}} to \math{T}.
\\
\math{\bullet}& There is no trusted third party and the peer-to-peer
communication is small (sub-linear in $\sum_m n_m$).\\
\math{\bullet}&
For any document $x$, a coalition of parties with indices $\ell_1,\ldots,\ell_j$ that are
given only the communication transcript and their own databases \math{X^{(\ell_1)},\ldots,X^{(\ell_j)}}
cannot determine if $x\in X^{(m)}$,  where $m\not\in\{\ell_1,\ldots,\ell_j\}$, 
in polynomial time.\\
\end{tabular}}
}
}}}
\caption{PD-NMF Problem Formulation.\label{fig:prob-statement}}
\end{figure}

\begin{remark}[On the exactness requirement] Our treatment crucially differs from DP-based approaches to private distributed learning in that we insist on computing the exact solution $T$ to the NMF problem. 
%
Exactness is important. Since each of the $M$ parties already has local data $X\m$, they can already compute a local NMF of $X\m$ and get an estimate $T\m$ of $T$.
The reason they are participating in a distributed computation it to {\sl improve their local estimate} $T\m$ to the $T$ that results from the full $X$.
Noising the output in the process would corrupt the estimate of $T$,
which defeats the purpose of participating in the distributed protocol.
In practice the deterioration is drastic.
\end{remark}

Toward constructing such an algorithm,  we adapt an existing versatile and efficient (but non-private and
non-distributed) NMF algorithm~\cite{honmf} to the distributed scenario. Our adaptation is carefully crafted so that parties running it need to {\sl only} communicate sums of locally computed vectors; these sums are distributively computed by means of a very light-weigh cryptographic (MPC) primitive,
namely `secure multiparty sum,' denoted as \ssum. \ssum   provides provable (full) privacy guarantees,
hiding each party's contribution to the sum,
at (small) constant or no communication overhead.
We call our adapted algorithm Private Distributed NMF (PD-NMF).
PD-NMF's communication cost is $O(Mdk)$ per iteration in the peer-to-peer communication model, and {\sl is independent of the database size $n$.} 
As a consequence it is possible to compute the distributed NMF using even less communication than what would be required to aggregate all data to a central server. We formally prove that given the same initialization, PD-NMF converges to {\it exactly} the same solution as centralized NMF. Furthermore, we develop a new
private distributed
initialization for our distributed NMF algorithm which
ensembles locally computed NMF models so that the global
PD-NMF converges quickly and to a good solution.

PD-NMF communicates the sums of intermediate results. Intuitively, due to the statistical properties of sums---i.e. that a single random term completely randomizes the entire sum---if the data itself is sufficiently unpredictable then the aggregate sum does not reveal considerable information on any individual record. In order to quantify the leakage from revealing the intermediate sums, we use an idea inspired by {\sl distributional differential privacy} (DDP), also known as {\sl noiseless DP~\cite{balcan2012distributed,bassily2013coupled,bassily2016typical}.} Informally, in such a notion of privacy we get a similar guarantee as in DP, i.e., indistinguishability of neighboring databases, but this guarantee is realized
due to the entropy of the data themselves. More concretely, an output mechanism is rendered private if for any individual record, the output of the mechanism on the databases with and without this record---using the terminology of (D)DP, {\sl on neighboring databases} is indistinguishable (i.e., the output distributions are close). 

The reason why the above implies an intuitive notion of privacy is similar to DP: An adversary observing the output cannot determine whether or not any given record is in the database any better than he can distinguish the two neighboring databases (with and without this record). This notion of privacy is sensitive to both the original data distribution---since the indistinguishability stems from the entropy of the data instead of external noise---and to prior information---since priors change the actual distribution of the data from the point of view of an attacker. Both these quantities are parameterizing the privacy. In this work, we aim to define privacy against other participants in the (distributed) learning protocol. Therefore, we will assume the data of those participants as the side-information (i.e., the prior) of the adversary trying to distinguish
between possible neighbouring databases of the victim.

A major obstacle, however, is actually estimating the above distinguishing advantage---or the distance in the corresponding distributions which is an upper bound to the advantage of any distinguisher. In the simple one-shot mechanism considered in the strawman examples of~\cite{balcan2012distributed,bassily2013coupled,bassily2016typical}, one can analytically compute the corresponding (posterior) distributions and therefore directly calculate their distance. This is unfortunately not the case in distributed learning applications, as even if we start with very clean initial data distributions---e.g., uniform distribution---in each iteration the input to the mechanism is updated as a (non-linear and often convoluted) function of the state of the previous iteration, which make the  analytical calculation of the distribution of  the output of later iterations infeasible. 

To overcome the above and adequately estimate the privacy loss incurred by an adversary learning the outputs of intermediate iterations of the learning algorithm, we propose a new {\sl experimentally measurable} notion of (distributional differential) privacy. 
%
We call this \textsl{Kolmogorov-Smirnov Distributional Privacy (KSDP)}. KSDP is similar in spirit to DDP  but uses the KS hypothesis testing method to define a notion of similarity (i.e. distance) between distributions. Loosely speaking, under KSDP, we preserve privacy if an adversary can not statistically distinguish between
PD-NMF being run on a database that contains a particular document $x$ compared to a when the database that doesn't have $x$. We stress again that the privacy guarantees offered by DDP (hence also by KSDP) are weaker than  DP. Indeed, DP ensures that the privacy of each individual record is protected irrespective of the prior information of the observer, whereas KSDP relies on the records in the database being sufficiently random; 
however, this is the inevitable
price one must pay
to ensure that the learning output is not corrupted by adding external noise,
as would be the case in a DP-line mechanism.

Last but not least, we demonstrate PD-NMF for topic modeling, and recommender systems
tasks~\cite{ding2008equivalence,salakhutdinov2007probabilistic}
on four well-known datasets, where our distributed initialization alone
matches the model quality of the centralized solution, and the output
satisfies our notion of privacy, i.e., KSDP.

\subsubsection{Overview of Private Distributed SVD}
We next apply our methodology to derive a protocol for private, distributed singular value decomposition (PD-SVD).
The SVD problem is defined as follows: Let $X$ be an $n\times d$ real-valued matrix (\math{n\ge d}).
The singular value decomposition (SVD) of \math{X} is the factorization
$
X = U\Sigma V^T,
$
where $\Sigma\in\R^{d\times d}$ is diagonal and
$U\in\R^{n\times d}, V\in\R^{d\times d}$
are orthogonal.
The \termdef{$k$-truncated SVD} of $X$ is defined as
\mandc{
X_k=U_k\Sigma_k V^T_k= \sum_{t=1}^k \Sigma_{tt} U_{:t}V_{:t}^T,
}
where
$U_k\in\R^{n\times k}$ contains the top-\math{k}
left singular vectors as its
columns $\{U_{:t}\}_{t=1}^k$,
$V\in\R^{d\times k}$ contains the top-\math{k} right singular vectors
as its columns $\{V_{:t}\}_{t=1}^k$,
and $\Sigma\in\R^{k\times k}$ is diagonal with diagonal entries
$\{\Sigma_{tt}\}_{t=1}^k$ containing the top-\math{k} singular values.
In a typical machine learning setting, each of the $n$ rows of
$X$ is a data point (e.g., documents) and each of the
$d$ columns is a feature (e.g., term in a document).
In a document-term setting, 
the columns of $V_k$ can be interpretted as topics,
i.e., combinations of terms, and each entry in a row of $U_k\Sigma_k$
is the amount of each topic in the corresponding
document~\cite{deerwester1990indexing}.
 

The factorization \math{X_k=U_k\Sigma_k V_k^T} is 
important in machine
learning algorithms (e.g., feature extraction, spectral
clustering, regression, topic discovery, etc.),
because it is an optimal orthogonal decomposition of the data:
$||X-X_k||\le||X-\hat X||$ for any rank-$k$ matrix
$\hat X$ and unitarily invariant norm (e.g., spectral or Frobenius).

As with PD-NMF, in the private distributed version of the problem the parties hold their own private data and the goal is to compute the SVD on the union of this data. The formal problem statement for private distributed SVD
can be found in Figure~\ref{fig:prob-statementSVD}. 

\begin{figure}[ht!]
\centerline{\fboxsep5pt\fbox{\parbox{0.8\columnwidth}{
\textbf{PD-SVD: Private Distributed SVD.}
{$M$ mutually distrustful parties each have an $n_m\times d$ matrix
  $X^{(m)}$ of $n_m$ rows over \math{d} features, sampled from an underlying
  distribution $\mathcal D$. The parties agree on the $d$ features,
  and their order. 
Given a rank $k$: 
each party must compute the $V_k$ such that:
\\[2pt]
{\tabcolsep1pt
\begin{tabular}{p{0.02\linewidth}p{0.95\linewidth}}
\math{\bullet}&
$V_k$ contains the optimal `topics'  (as defined above) for the full data $X$, where $X$ is computed by stacking the rows of $X^{(m)}, m\in\{1,\ldots, M\}$ into a single matrix. 
%
%
\\
\math{\bullet}& There is no trusted third party and the peer-to-peer
communication is small (sub-linear in $\sum_m n_m$).\\
\math{\bullet}&
For any document $x$, a coalition of parties with indices $\ell_1,\ldots,\ell_j$ that are
given only the communication transcript and their own databases \math{X^{(\ell_1)},\ldots,X^{(\ell_j)}}
cannot determine if $x\in X^{(m)}$,  where $m\not\in\{\ell_1,\ldots,\ell_j\}$, 
in polynomial time.\\
\end{tabular}}
}
}}}
\caption{PD-SVD Problem Formulation.\label{fig:prob-statementSVD}}
\end{figure}

Our starting point towards PD-SVD is
a simple block power-iteration method for 
jointly finding the singular values/vectors of a matrix.
There are more sophisticated methods (see for example \cite{Parlett1998}), however
our focus is not on robustness and efficiency of the SVD-algorithm, but rather the
ability to recover the centralized solution privately in a distributed
setting.
Since we are
to compute the row basis
\math{V_k}, we can reduce the problem to the symmetric covariance
matrix,
\math{X^TX=V\Sigma^2V^T}. It is convenient that
\math{X^TX} is a sum over individual party covariance matrices,
\mandc{X^TX=\sum_{m=1}^M{X^{(m)}}^TX^{(m)}.}
  This means that any linear operations on   \math{X^TX} are a sum of those
  same linear operations over the local covariance matrices.
  One additional benefit of using covariance matrices is that the communication
  complexities depend on \math{d}, not \math{n=\sum_m n_m}.
  The downside of using the covariance matrix is the the condition
  number is squared, so this affects numerical stability of all algorithms,
  but that is not our main focus and we will assume that
  the data matrices are well behaved.
  The
  cryprographic
  challenges for SVD is that we only want to reveal \math{V_k}, and not
  \math{\Sigma_k} (the eigenvalues) or \math{U_k} (the left singular vectors),
  because
  those would contain much more information, and essentially reveal the full data, or a close approximation to it.
  As a result, \ssum alone will not be enough to privately compute
  only \math{V_k} (and hide \math{\Sigma_k}).

We note in passing that the security community has studied algorithms for the distributed computation of the exact SVD under a variety of security constraints.
However a consistent issue is that in addition to $V_k$, $\Sigma_k$ is typically revealed.
In \cite{SEC:SEC1501,DBLP:journals/corr/ChenLZ17a} the methods essentially reduce to sharing $X^TX=V\Sigma^2 V^T$ securely, and as a result each party learns $\Sigma$.
In \cite{4812517} a method is proposed based on the QR decomposition that allows  parties to only share $\Sigma$ if they choose to, however it is only developed for $M=2$ parties.
In \cite{youdao2010p4p}, the SVD is solved using a method similar to ours, but once again $\Sigma$ is revealed.\footnote{The latter work has many other interesting ideas. For example random projections are used as a method to build zero knowledge proofs of the fact parties participating in the distributed SVD computation are using the same input matrix round-to-round. These are enhancements that could be added to our algorithm and their effectiveness is left as future research.}
Finally, all of the above works have not attempted to address the document privacy issue.

 To avoid revealing $\Sigma_k$ in the power iterations, we will use another lightweight cryptographic (MPC) module which we term \textit{normalized secure sum}, denoted as \nssum. (In fact, our algorithm will use both \nssum and the original \ssum.) \nssum is similar to \ssum, i.e., receives from each party a vector $\mbf x_m$, but instead of simply securely computing and outputting the sum $\mbf s=\sum_{m=1}^M \mbf x_m$ of the vectors, it computes and output the sum normalized by their L2 norm, i.e.,  $$\nssum(\{\mbf x_m\}_{m=1}^M)=\frac{\mbf s}{||\mbf s||_2}.$$
 We remark that although sightly heavier, in terms of communication and computation, than the very lightweight \ssum, \nssum is still practically computable and in any case the overhead compared to the learning algorithm computation is very low as we demonstrate in our experiments. In fact, with further cryptographic optimizations this overhead can be brought down to hardly noticeable (cf.~\cite{NDSS:D0Z15} and references therein.)  Most importantly, by only communicating over the lightweight cryptographic primitives we ensure that the
 communication cost of our algorithm does not depend on the number of rows $n$.

 The privacy of our PD-SVD algorithm is argued analogously to PD-NMF, i.e., we use KSDP to estimate the distance between the the distributions of (vectors of) intermediate outputs for \ssum and \nssum
 for neighboring data distributions,
 i.e., for data matrices with and without any one individual record.  

\subsubsection{Summary of Contributions}
\begin{itemize}
\item We propose a shift from enclosing distributed learning algorithms
  inside costly MPC frameworks to modifying learning algorithms to
  use only lightweight cryptographic protocols (PD-ML-Lite).
\item For two landmark machine learning problems (NMF and SVD)
  we show that 
  two extremely lightweight primitives suffice, \ssum and \nssum.
  Using just these two primitives we guarantee recovering the
  centralized solution.
\item We introduce Kolmogorov-Smirnov Distributional Privacy (KSDP) as an empirically estimable alternative to noiseless (aka distributional) differential privacy to measure the privacy leaked in both the intermediate outputs of \ssum,
  \nssum and the final outcome of the learning. Privacy is preserved through
  the entropy in the data distribution and the aggregate nature of the
  intermediate outputs.
  Experiments with real datasets shows that negligible privacy is leaked,
  a small price to pay in privacy, in return for the optimal centralized
  learning outcome.
\item  We show experimental results in
  Section~\ref{sec:experimental_results_both}, where
  we showcase the performance, accuracy, and privacy of our PD-NMF and
  PD-SVD algorithm in classical applications of NMF and SVD
  (topic modeling,
  recommender systems, principal component regression and
  low rank approximation). Privacy is essentially preserved with significant uplift in the learning
  outcome. In contrast, we show results for making the algorithms differentially
  private through addition of noise: privacy is guaranteed but the
  learning outcome is drastically deteriorated to the point where
  the local learning outcome becomes measurably better than the
  differentially private distributed outcome. In such a case, there is
  zero upside in participating in the distributed protocol.
\end{itemize}

\subsection{Notation.}
\math{X} is the row-concatenation of the individual matrices
\math{X^{(m)}}. For a matrix \math{A}:
\math{A_{i:}} is the \math{i}th row;
\math{A_{:j}} is the \math{j}th column;
\math{\norm{A}_{1,2}} are the \math{\ell_1} and \math{\ell_2} entry-wise norms
and \math{\norm{A}_F} is the Frobenius norm equal to the
\math{\ell_2} entry-wise norm (we use
\math{\norm{\cdot}_2} for vectors and \math{\norm{\cdot}_F} for matrices);
\math{[A]_+} projects \math{A} onto the non-negative orthant
by zeroing its negative entries.
$\mathbf 1_d$ is the \math{d}-dimensional vector of $1$s.

\remove{
The NMF of a matrix is a important building block for various machine learning algorithms. The privacy guarantee of our protocol makes application of NMF to solve problems topic modeling \citep{ding2008equivalence} (when rows correspond to documents) and recommender systems \citep{salakhutdinov2007probabilistic} (when rows correspond to a person's usage history) particularly interesting. Example current or potential use cases might be:
\begin{itemize}
\item A distributed peer-to-peer information retrieval system based on topic models. \cite{tang2003peer,yi2009comparative} For example government agencies could query each other's databases for topics before going through full legal procedures to get full text. Or a distributed social network could keep track of trending topics \citep{yamron1999topic} without revealing what posts are stored on each node.
\item Cooperative business-to-consumer recommender systems \cite{5477138}. For example businesses that don't directly compete can help each other better understand their customer's shopping habits (e.g. the ``Plenti'' rewards card). Or a free music streaming service could cooperate with a pay-to-own service like the iTunes Music Store to give each others' customers better recommendations.
\item Compartmentalizing data-security so that a breach in one location doesn't affect all data. \cite{anderson2008security} This is relevant for applications as diverse as sensor networks in environments with eavesdroppers \cite{chan2003security}, or cross-jurisdictional learning where the transfer of user data between a business' own servers in different jurisdictions is governed by privacy laws like the EU-US ``Privacy Shield.''
\item Even if privacy isn't important, since our protocol is communication-optimal, as a distributed dataset grows, our algorithm will be able to compute the NMF faster than it would take to just transfer the data to one central location.
\end{itemize}

In order to clearly relate our work to the rich and expanding literature in our field, we need to first state the problem we solve precisely.

\fbox{\parbox{0.98\textwidth}{
\textbf{Problem Setting: Private Distributed NMF.}
Suppose there are $M$ mutually distrustful parties, and a trusted third party is not available. Each party holds a non-negative matrix $X^{(m)}$ of $n_m$ rows over a agreed-upon set of $d$ features. Suppose there is a agreed-upon rank $k$. Each party should find the same non-negative $k\times d$ global basis matrix $T$ such that the Frobenius reconstruction error
\begin{equation}
\frac{1}{2}\sum_{m=1}^M ||X^{(m)}-W^{(m)}T||_F^2
\end{equation}
is minimized when each party fits a local non-negative $n_m\times k$ representation matrix $W^{(m)}$. Importantly, the protocol should not leak to any party $p$ whether or not $\mathbf x \in X^{(m)}$ for $m\neq p$.
}}
}

\subsection{Organization of the Paper}
In Section~\ref{sec:algorithm} we give the details of our PD-NMF algorithm and prove its exact convergence properties. In Section~\ref{sec:algorithmsvd} we describe our PD-SVD protocol and analyze its accuracy. Our new privacy definition (KSDP) which is used for the privacy analysis of both PD-NMF and PD-SVD) is then given in Section~\ref{sec:privacy_analysis}. Finally in Section~\ref{sec:experimental_results_both} we provide our experimental results for both PD-NMF and PD-SVD. For completeness we have also included a clearly marked and structured appendix, with details on our claims and experiments, which is referred throughout as appropriately. 

%% file: algorithm.tex
\section{Private Distributed NMF (PD-NMF)}
\label{sec:algorithm}

\newcommand{\reg}{\text{reg}}
\newcommand{\nnsvd}{\text{nnsvd}}
\newcommand{\iter}{\textsc{PD-NMF-Iter}\xspace}
\newcommand{\init}{\textsc{PD-NMF-Init}\xspace}
\newcommand{\pd}{\textsc{PD-NMF}\xspace}

{
\makeatletter
\def\thm@space@setup{\thm@preskip=5pt
\thm@postskip=5pt}
\makeatother

\makeatletter
\newenvironment{pf}[1][\proofname]{\par
  \pushQED{\qed}%
  \normalfont \topsep0\p@\relax
  \trivlist
  \item[\hskip\labelsep\itshape
  #1\@addpunct{.}]\ignorespaces
}{%
  \popQED\endtrivlist\@endpefalse
}
\makeatother

Given a rank $k\in\mathbf N$, the objective is to find non-negative matrices \math{W,T} that (locally) minimize
\mldc{
\E(W,T)=0.5||X-WT||_F^2
+\reg(W,T),\label{eq:nmf_obj}
}
where $W$ is $n\times k$, $T$ is $k\times d$ and
the regularization term is given by 
\mldc{
\reg(W,T)=\alpha||T||_{1}
+{\textstyle\frac12}\beta||T||_F^2
+\gamma||W||_{1}
+{\textstyle\frac12}\delta ||W||_F^2. \label{eq:nmf_req}
}
(\math{\alpha,\beta,\gamma,\delta} are regularization hyperparameters.)
When either $T$ or $W$ is fixed, NMF reduces to non-negative least squares.
Hence, given $X$ and $T$ one can find an optimal $W$.
We can write~\r{eq:nmf_obj} as a sum of \math{k} rank-1 terms: 
Let residual \math{R_t} be the part of \math{X} to be explained by
topic \math{t},
\mldc{ R_t=X-\sum\limits_{l\neq t}^k W_{:l}T_{l:}
\qquad\text{for \math{t=1,\ldots,k}}.\label{eq:nmf_resid}}
The objective \math{\E} becomes
\mandc{
  \E(W,T)=\frac{1}{2k}
  \sum_{t=1}^k \norm{R_t - W_{:t}T_{t:}}_F^2+ \reg(W,T).
}
We now use a
projected gradient rank one residual method ( \textsc{RRI-NMF})
proposed in~\cite{honmf} to minimize~\math{\E}. This
method alternates between fixing $W_{:t}$ (resp. $T_{t:}$)
and optimizing for  $T_{t:}$ (resp. $W_{:t}$)
sequentially for each $t$ until a convergence.
The  iterative update of each topic $t$ is given
by~\cite{honmf}:
\mldc{
\begin{array}{l}
\displaystyle
W_{:t} \gets \frac{[R_tT_{t:}^T-\gamma\mathbf 1_n]_+}{||T_{t:}||_2^2+\delta}
\qquad\text{ for \math{t=1,\ldots,k}}; 
\\[14pt]
\displaystyle
T_{t:} \gets \frac{[W_{:t}^TR_t - \alpha \mathbf 1_d]_+}{||W_{:t}||_2^2+\beta}
\qquad\text{ for \math{t=1,\ldots,k}}; 
\\[14pt]
\displaystyle
T_{t:}\gets \argmin_{x\ge0, ||x||_1=1} ||x-T_{t:}||_2
\qquad\text{ for \math{j=1,\ldots,k}}.
\end{array}
\label{eq:nmf_update}
}
In the last step, ignoring the regularization, i.e. $\reg(W,T)=0$,
there is a diagonal scaling degeneracy in the factorization
because \math{WT=WD^{-1}DT}.
So we may assume without loss of generality that
the rows of \math{T} are normalized, i.e. \math{\norm{T_{j,:}}_1=1}.
This projection onto the simplex can be done efficiently~\cite{condat2016fast}.
In the presence of regularization this constraint can be retained or dropped depending on the application,
however simply projecting isn't correct in the presence of regularization.
In practice,
 \textsc{RRI-NMF} and other similar alternating non-negative least squares methods
\cite{lin2007projected,cichocki2007hierarchical} 
converge faster and to better solutions than the
original multiplicative algorithms for NMF~\cite{seung2001algorithms}.

The main point is that we 
solve Problem \ref{eq:nmf_obj} by 
decoupling the updates of $W$ and $T$. The update of $W$ is a sum of columns of the data $X$ (since $R$ is a function of $X$ by Equation \ref{eq:nmf_resid}) and can be done locally by each party. The update of $T$ depends on a sum of rows of the data $X$. Each party can evaluate their portion of the sum (given by the expression $(W_{:t}^{(m)})^TR_t^{(m)}$) on their $X^{(m)}$. Then, the parties can use the \ssum protocol (see below) to share these local sums so that everyone learns just the global sum. The observed output $W_{:t}^TR_t$ is the \emph{sum} of functions over
iid rows of $X$ which is not sensitive to specific rows;
 intuitively, this will ensure the privacy on \pd as \math{X} gets large.

\subsection{Secure Multiparty Sum Protocol (\ssum)}
The protocol \ssum is a standard lightweight cryptographic distributed peer-to-peer protocol (MPC) among $M$ parties $p_1,\ldots,p_M$.
It takes an input from each $p_m$ a vector $\vec{x}_m$  and outputs (to everyone) the sum $\textstyle\sum_{m\in M} \vec{x}_m$.  Any subset of parties only learns this sum and no additional information about the inputs of other parties. For completeness we have included details on  \ssum  in Appendix~\ref{app:smsp}. 

\subsection{\bf Private Distributed \textsc{RRI-NMF} Iterations (\iter)}
We use \ssum
to implement private distributed \textsc{RRI-NMF} iterations.
The term `iterations' means that we are making iterative improvements to an initial set of basis $T_0$.
We call this algorithm \iter, and its details are presented in Algorithm~\ref{alg:distributed_rri_nmf}.
Our full algorithm \pd consists of first running \init (presented later) to get an initial set of $T_0$, and then
use $T_0$ as the initial topics for \iter.
Figure~\ref{fig:pd} shows the workflow of PD-NMF.

\begin{figure}[hbtp]
\centerline{
  \scalebox{0.95}{
\begin{tikzpicture}[line width=1pt,>=latex,inner sep=2pt,x=1.75cm,y=0.95cm]
\draw[line width=1pt,dashed](-0.6,-1.7)rectangle(4.5,1.8);
\draw[line width=1pt](-0.7,-1.9)rectangle(9,1.9);
\foreach\m[count=\i]in{1,...,3}{
\node[](X\m)at(0.0,1.9-\m){\math{X^{(\m)}}};
\node[](T\m)at(1.75,0.8*2-0.8*\m){\math{T^{(\m)}}};
\node[](Y\m)at(5.75,1.9-\m){\math{X^{(\m)}}};
\path(X\m)edge[->]node[sloped,above=0pt,scale=0.75]{NMF}(T\m);
}
\foreach\x[count=\i]in{-1,1}{
\draw[dotted](-0.5,0.5*\x)--(0.5,0.5*\x);
\draw[dotted](1.25,0.7*0.5*\x)--(2.25,0.7*0.5*\x);
\draw[dotted](5.25,0.5*\x)--(6.25,0.5*\x);
}
\node[](Tinit)at(4.25,0){\math{T_{\text{init}}}};
\node[](Tfinal)at(8.5,0){\math{T^G_\text{final}}};
\draw($(X3)+(-0.5,-0.35)$)rectangle($(X1)+(0.5,0.35)$);
\draw($(Y3)+(-0.5,-0.35)$)rectangle($(Y1)+(0.5,0.35)$);
\draw($(T3)+(-0.5,-0.35)$)rectangle($(T1)+(0.5,0.35)$);
\path
(2.25,0)edge[->]node[pos=0.5, above=1pt, rotate=0]{\iter}(Tinit)
(2.25,0)edge[->]node[below=4pt,scale=0.8]{\begin{tabular}{r}Weigh topics,\\and start from\\random $T_0.$\end{tabular}}(Tinit)
(6.25,0)edge[->]node[pos=0.5, above=1pt, rotate=0]{\iter}(Tfinal)
(6.25,0)edge[->]node[pos=0.5,below=0pt]{\math{T_{\text{init}}}}(Tfinal)
;
\draw[<-](5.25,0)--(Tinit);
\node[red,font=\small,anchor=east]at(4.475,1.55){\init};
\node[red,font=\large,anchor=east]at(7.9,1.6){\textbf{\pd}};
\end{tikzpicture}}}
\label{fig:pd}
\caption{\pd.}
\end{figure}

\begin{figure}[t]
  \tabcolsep0pt
\begin{tabular}{p{0.5\textwidth}p{0.5\textwidth}}
\begin{minipage}[t]{0.45\textwidth}
  \begin{algorithm}[H]
\caption{\iter: Private Distributed \textsc{ \textsc{RRI-NMF}}, party \math{m}'s view.}
\floatname{algorithm}{Procedure}
\begin{algorithmic}[1]

\renewcommand{\algorithmicrequire}{\textbf{Input:}}
\renewcommand{\algorithmicensure}{\textbf{Output:}}
\Require Local data $X^{(m)}$, initial topics $T$, parameters $\alpha,\beta,\gamma,\delta$, \SMSP
\Ensure Global topics $T^G$ 
\State {\bf repeat} until convergence
\For{$t\in\{0,1,\ldots,k\}$}
\State $R_t^{(m)} \gets X^{(m)}-W^{(m)}T+W_{:t}^{(m)}T_{t:}$ 
\State $\displaystyle W_{:t}^{(m)} \gets \frac{[R_t^{(m)}T_{t:}^T-\gamma\mathbf 1_{n_m}]_+}{||T_{t:}||_2^2+\delta}$
\State \math{num\gets\ssum(W_{:t}^{(m)T}R_t^{(m)}
)}
\State \math{den\gets\ssum(||W_{:t}^{(m)}||_2^2,
)}
\State $\displaystyle T_{t:} \gets \frac{\left[num-\alpha\mathbf1_d\right]_+}{den+\beta}$
\State $T_{t:}\gets \argmin\limits_{x\ge 0, ||x||_1=1} ||x-T_{t:}||_2$
\EndFor 
\end{algorithmic}\label{alg:distributed_rri_nmf}
  \end{algorithm}
  \end{minipage}
&
\hfill
\begin{minipage}[t]{0.45\textwidth}
  \begin{algorithm}[H]
\caption{\init: initialization by merging local topics, party $m$'s view.}
\label{alg:init}
\floatname{algorithm}{Procedure}
\begin{algorithmic}[1]
\renewcommand{\algorithmicrequire}{\textbf{Input:}}
\renewcommand{\algorithmicensure}{\textbf{Output:}}
\Require Local data $X^{(m)}$, number of topics~$k$, \iter, \ssum
\Ensure Initial global topics $T_\text{init}$
\State Initialize $T_\text{loc} \gets$ \verb+nnsvd+($X^{(m)},k$).
\State $W^{(m)}, T^{(m)} \gets \textsc{RRI-NMF}(X^{(m)},\,k,\,T_\text{loc})$.
\State Initialize $V^{(m)}\gets 0^{k\times k}$
\State Set $V^{(m)}_{jj} \gets \sqrt{\sum_{i=1}^n (W_{ij}^{(m)})^2}$.%
\State Scale $\hat T^{(m)} \gets V^{(m)}T^{(m)}$.
\State Sample $T_0^{(m)} \sim U[0,1]^{k\times d}$.
\State Compute $T_0 \gets \ssum(\frac{1}{M} T_0^{(m)})$.
\State Normalize $T_0$'s rows to sum to 1.
\State $T_\text{init} \gets $ \iter$(\hat T^{(m)}; T_0)$.
\end{algorithmic}
\end{algorithm}
\end{minipage}
\end{tabular}
\end{figure}

We give a simple but essential  theorem on the learning outcome
of \iter (see Appendix~\ref{app:omitted_proofs} for the proof).
This theorem says that our distributed algorithm mimics the
centralized algorithm.
\begin{theorem}\label{thm:pdnmfiter}
Starting from the same initial topics \math{T_0},
\iter (Algorithm \ref{alg:distributed_rri_nmf}) 
converges to the same solution $T$ as the centralized \textsc{RRI-NMF}.
\end{theorem}

{\bf Initializing $T$ with \init.}
As with any iterative algorithm, \textsc{ \textsc{RRI-NMF}}
must be initialized. There are different ways to initializate
each with their pros and cons.
For non-private NMF there are several state-of-the-art initialization algorithms~\cite{boutsidis2008svd,kumar2012fast,gillis2014successive}
that are comparable in terms (a) computational complexity (b) solution quality after $\tau$ iterations.
SVD-based initialization (\verb+nnsvd+) of~\cite{boutsidis2008svd}
produces $T_0$ from a row-basis for the top singular  subspaces of~\math{X}.
While convergence from $T_0$ is quick, \verb+nnsvd+ needs access to $X$
or a private distributed SVD algorithm - we do give a private distributed
SVD algorithm later, so that could be used here.
A much simpler initialization (\verb+random+)
chooses each entry of $W,T$
independently from $U[0,1]$ and
is private by construction, but converges slowly.
In our work flow, we propose a new
distributed initialization algorithm.
\init.
Our initialization algorithm uses \iter as a subroutine. Hence,
the privacy of \init depends on the privacy of \iter.
The quality of our new initialization is
comparable to \verb+nnsvd+ in quality
(Section \ref{sec:experimental_results}).
The key idea is that since $X^{(m)}\approx W^{(m)}T^{(m)}$,
appropriately weighted rows of $T^{(m)}$
are good representatives for rows of $X^{(m)}$.
We treat a weighted $T$, denoted $\hat T$, as a
pseudo-document corpus and run \iter on these pseudo-corpora, initialized with
\verb+random+. The output is
 $T_\text{init}$, which is the initialization
for a second round of \iter run on the full document corpus $X$.
This entire process is \pd (see Figure \ref{fig:pd}).
We give the details for \init in Algorithm \ref{alg:init}.

The next theorem
has two parts. The first is on the quality of \math{T_\text{init}}.
 Let
 $\E_Y(Z)=\min_{W\geq 0} 0.5||Y-WZ||_F^2$ be the lowest possible reconstruction error for $Y$ given basis $Z$.
 The choice of the
 weights for the rows of $\hat T$ (step 4 of \init)
 ensure that we minimize 
\math{\E_{\hat T}(Z)} w.r.t. \math{Z}) to obtain
\math{T_\text{init}}. It turns out that this minimizes an upper bound on the
true objective that we wish to minimize
\math{\E_{X}(Z)}, and this upper bound is tight
when the rows of $\hat T$ are mutually orthogonal. This means that our
initial topics from which to run \iter are good, as are the topics initialized
by SVD. This explains why our new initialization is competitive with the
state of the art. The advantage or our initialization is that it is easy
to make private: it suffices to make \iter private. The theorem summarizes
these
facts (see Appendix~\ref{app:omitted_proofs} for a proof). 
\begin{theorem}[Quality and privacy of initialization]\label{theorem:upper}
  \begin{inparaenum}[(i)]
    \item
  Our initialization algorithm \iter (Algorithm \ref{alg:init})
  is private if \iter is private.
\item
  The output of \iter is a set of topics $T_\text{init}$ that minimizes,
  with respect to \math{Z},
  \math{\E_{\hat T}(Z)}.
  Further, \math{\E_{\hat T}(Z)} is an upper bound on \math{\E_X(Z)}
  which is worst-case tight.
  \end{inparaenum}
\end{theorem}
Though we have not discussed privacy, the theorem says that our initialization
algorithm inherits the privacy guarantees that \iter provides when
run on \math{T}, whatever those be. In other words, privacy is not
leaked anywhere else in the algorithm.

%% file: algorithm_svd.tex
\section{Private Distributed SVD (PD-SVD)}
\label{sec:algorithmsvd}

We next turn to the algorithm for private distributed
singular value decomposition, PD-SVD.  Most methods for finding the
singular values/vectors of a matrix reduce the problem to finding
eigenvalues and eigenvectors
of a symmetric matrix~\cite{trefethen1997numerical,Parlett1998}.
We use the $d\times d$ symmetric matrix $S=X^TX$, so that the eigenvalue decomposition reveals the right-singular vectors $V_k$: 
\mandc{
\begin{aligned}
S &\approx V_k\Sigma_k U_k^TU_k\Sigma_k V_k^T\\
&= V_k\Sigma^2_k V^T_k.
\end{aligned}
}
In order to find the top $k$ eigenvectors we will use block power iterations,
starting with a random Gaussian
initialization. This (non-private) centralized
algorithm is summarized
in Algorithm~\ref{alg:non_private_power_iteration}. To ensure that
all parties start at the same initial state, the parties agree on a
pseudorandom number generator and broadcast the seed.  
The framework of the algorithm is similar to PD-NMF, so we give
only the high-level details. The basic iteration is
\mandc{
v_{t+1}=\frac{X^TX v_t}{\norm{X^TX v_t}}.}
Because \math{X^TX v_t=\sum_{m=1}^M{X^{(m)}}^TX^{(m)} v_t}, a naive approach would
ask each party to compute their share \math{{X^{(m)}}^TX^{(m)} v_t} and then the
sum can be securely shared using \ssum. This however will also share
\math{\norm{{X^{(m)}}^TX^{(m)} v_t}}, which are the eigenvalues in
\math{\Sigma_k}.
In order 
where to avoid revealing $\Sigma_k$ in the power iterations,
need to introduce a new 
cryptographic primitive that privately computes and shares
the normalized sum \math{{X^TX v_t}/{\norm{X^TX v_t}}}.
We call this cryptographic primitive \textit{normalized secure sum},
denoted as \nssum---this is the sum of input vectors divided by there \math{\ell_2}-norm.
The formal definition of this primitive can be found in Appendix~\ref{app:spdz}.

To build \nssum, we rely on a state-of-the-art MPC framework called SPDZ~\cite{cryptoeprint:2010:514,cryptoeprint:2011:535,cryptoeprint:2011:091,cryptoeprint:2017:1230}. SPDZ takes a procedural program as input and compiles it into an arithmetic circuit over a sufficiently large finite field.
This circuit is then evaluated in two phases by a `distributed virtual machine' run by the $M$ parties.
In the {\it offline} phase SPDZ generates pre-computed primitives which are then used in the {
\it online} phase to compute the compiled circuit in a cryptographically secure manner. The offline phase is computation agnostic, and can be performed independently of what the online phase will do; in our setting the $M$ parties could start the offline phase before they've even begun collecting their $X\m$s.

 We use SPDZ because of its efficient online phase: the computation and communication complexity is linear in the number $M$ of parties,  the size of the circuit which is output by the compiler, and the input size $d$. The actual running time of compiled circuit depends on both the procedural code as well as the SPDZ compiler, therefore we evaluate this experimentally. This evaluation, as well as implementation details and further discussion are available in Appendix \ref{app:spdz}.
In addition to \nssum we make use of the protocol  \ssum from the previous section (recall that this protocol computes the non-normalized sum of its inputs).

The matrix-vector
multiplication and rescaling steps in the centralized
SVD (Algorithm~\ref{alg:non_private_power_iteration})
are simultaneously handled by \nssum in our private-distributed version of the
algorithm, denoted as \pdsvd (Algorithm \ref{alg:pdsvd}).
The following theorem states that \pdsvd computes the same output as the centralized SVD algorithm (we refer to Appendix~\ref{app:omitted_proofs} for a proof.) 

\begin{theorem}\label{thm:svd}The output of \pdsvd matches the centralized non-private SVD exactly up to a user-specified numerical precision, assuming they
are both initialized from the same seed.
\end{theorem}
\begin{table}[t]
\tabcolsep0pt
\begin{tabular}{p{0.5\textwidth}p{0.5\textwidth}}
\begin{minipage}[t]{0.46\textwidth}
\begin{algorithm}[H]
\begin{algorithmic}[1]
\caption{Non-private block power iteration.}
\label{alg:non_private_power_iteration}
\floatname{algorithm}{Procedure}
\Require $d\times d$ matrix $S$, truncation rank $k\in\N$, number of
iterations$\tau\in\N$.
\Ensure $\hat V_k$: estimate of top $k$ eigenvectors of $S$.
\State $\hat V\gets \mathcal N(0,\frac{1}{d})^{d \times k}$
\For{$t\in\{1,2,\ldots,\tau\}$:}
\State $\hat V \gets S\hat V$
\State $\hat V \gets \text{Orthonormalize }(\hat V)$
\EndFor
\State Output $\hat V$ as an estimate of $V_k$. 
\end{algorithmic}
\end{algorithm}
\end{minipage}
&
\hfill
\begin{minipage}[t]{0.46\textwidth}
\begin{algorithm}[H]
\begin{algorithmic}[1]
\caption{PD-SVD, party $m$'s view.}
\label{alg:pdsvd}
\floatname{algorithm}{Procedure}
\Require $n_m\times d$ matrix $X^{(m)}$,  truncation rank $k\in\N$,
 number of iterations$\tau\in\N$, number of parties $M$, \ssum and \nssum.
\Ensure $\hat V_k$: estimate of top $k$ eigenvectors of $S$.
\State $S^{(m)}\gets \left(X^{(m)}\right)^TX^{(m)}$
\State $\hat V^{(m)} \gets \mathcal N(0,\frac{1}{\sqrt M d})^{d\times k}$
\State $\hat V \gets \ssum(\hat V^{(m)})$
\For{$t\in\{1,2,\ldots,\tau\}$:}
\For{$i\in\{1,2,\ldots, k\}$:}
   \State $\hat V_{:i} \gets \nssum{}(S^{(m)}\hat V_{:i})$
\EndFor
\State $\hat V \gets \text{Orthonormalize }(\hat V)$
\EndFor
\end{algorithmic}
\end{algorithm}
\end{minipage}
\end{tabular}
\end{table}

A common application of the SVD in machine learning is to compute the principal components of a set of points represented as rows of a matrix. \pdsvd can be extended to perform the PCA by having each party center its database, details are in Appendix \ref{app:pca}.

%% file: privacy_analysis.tex
\section{Privacy Analysis and KSDP}
\label{sec:privacy_analysis}

We next discuss the privacy guarantees of our algorithms/protocols.  To better understand privacy in our context it is useful to define the concept of an {\it observable}.
\mandc{
\fboxsep5pt\fbox{\parbox{0.91\columnwidth}{
\textbf{Observable.}
Suppose a distributed mechanism $\mathcal A$ is executed among several parties. The party of interest uses database  $X$;  the concatenation of the remaining parties' databases is $Y$. Then $\mathcal O_{\mathcal A, X, Y}$ is the set of objects that a coalition of adversaries observes from the communication transcript of $\mathcal A$.
}}}
In a nutshell, privacy of our protocol would
require that the concatenation of the observables of all iterations
does not reveal substantial information on any individual row in
\math{X} to the adversary. Our goal is to measure the 
leakage of information from the observables.
In principle, we may exclude from this leakage any information
leaked by the outcome of the learning itself, since the goal is for
all parties to obtain the centralized learning outcome. However, in deciding
whether to participate in the distributed learning protocol, a party may
wish to know the information leaked from the outcome of the learning itself.
Our method for measuring privacy can accomodate any set of observables,
including
the outcome of the learning.

Intuitively, our protocol satisfies
such a notions of privacy, because the adversary only sees (amortized) sums
which, assuming sufficient uncertainty in the honestly held data,
is blinded by the randomness of the  contributions from honest parties.
More formally,
a mechanism $\mathcal A$ preserves privacy of any document $x$
if it's output can be simulated by an $x$-oblivious
mechanism $\mathcal B$. So,
$ \mathcal  B$,
without access to $x$, can emit {\it all} observables $\mathcal O_{\mathcal A, X,Y}$ with a comparable probability distribution to $\mathcal A$.
Our definition 
is similar to the well-accepted Distributional Differential Privacy (DDP)~\cite{bassily2013coupled,groce2014new}, which implies
privacy of individual rows of each party's database~\cite{kairouz2016fundamental}. 
 DDP is similar to DP, but uses data entropy rather than external noise to guarantee privacy.
 Mathematically, mechanism $\mathcal A$ is  {\sl distributionally differentially private} (without auxiliary information) for dataset $X$ drawn from $\mathcal U$,
 and all individual rows $x\in X$ if for some mechanism $\mathcal B$,
\mandc{\Prob[\mathcal A(X)] \approx_{\epsilon, \delta} \Prob[\mathcal B(X\setminus\{x\})],
}
where for probability density functions $f,g$
over sample space $\Omega$ with sigma-algebra $\mathcal S \subseteq 2^\Omega$,
$f\approx_{\epsilon, \delta}g$ if, \math{\forall \omega \in \mathcal S},
\mldc{
\begin{aligned}
f(\omega) &\leq e^\epsilon g(\omega)+\delta \hspace*{1ex}
\text{ and } \hspace*{1ex} g(\omega) \leq e^\epsilon f(\omega)+\delta.
\end{aligned}
\label{eq:ddp}
}
Adding noise to achieve DP~\cite{dwork2014algorithmic,blum2005practical,wang2015privacy} satisfies \r{eq:ddp}, but is invalidated as it deteriorates
learning accuracy.\footnote{Theoretical results for DP (e.g. \citet{balcan2012distributed})
only apply to simple mechanisms. Composition of these simple mechanisms needs to be examined case-by-case
(e.g., in one-party Differentially Private NMF, \citet{FastDPMF}
incurr a 19\% loss in learning quality
when strict DP is satisfied even for $\epsilon=0.25$).
In order to improve the learning outcome,
\cite{FastDPMF} propose violating privacy
for atypical users (which lets them add less noise to the mechanism).
In the $M$-party setting due to a possible difference attack at successive
iterations, each party must add noise to all observables they emit in
every iteration~\cite{rajkumar2012differentially, hua2015differentially}.
The empirical
impact is a disaster.}
Having invalidated DP as a viable privacy model for distributed learning,
the next best thing to argue our protocol's privacy would be to theoretically
compute DDP. This is 
intractable for complex nonlinear iterative algorithms
(even
starting from friendly distributions) since \r{eq:ddp} must
hold at every iteration.
We take a different,
experimentally validatable approach to DDP suitable for
estimating privacy of (noisless) mechanisms.
We
introduce {\sl Kolmogorov-Smirnov (distributional) differential privacy},
KSDP, which may be of independent interest for
situations where DDP is not-computable and noising the outcome to guarantee
privacy is not a viable option.

Intuitively, the adversary tries to determine from the (high-dimensional)
observable whether \math{x} is in the victim's database. A standard
result is that there is a sufficient statistic
\math{\sigma \colon \mathcal O \times x \to \R}
with maximum discriminative power.
For such a statistic $\sigma$ and a document $x$,
let $f$ be the distribution of $\sigma$ with observables $\mathcal O_{\mathcal A, X,Y}$ and $g$ be the distribution of $\sigma$
with observables $\mathcal O_{\mathcal B,X\setminus\{x\},Y}$
(i.e. those emitted by a simulator $\mathcal B$ which doesn't
have access to $x$).
Equation~\r{eq:ddp} is one useful notion of similarity
between PDFs $f$ and $g$ (composability being one of the properties).
Equation \r{eq:ddp} is very difficult to satisfy, let alone prove.
It is also overly strict for protecting against polynomial-time adversaries.
We keep the spirit of distributional privacy,
but use a different measure of similarity.
Instead of working with  PDFs,
we propose measuring similarity between the corresponding CDFs $F$ and $G$
using the well-known Kolmogorov-Smirnov statistic
over an interval $[a,b]$,
\mandc{\text{KS}(F,G)\equiv\sup_{x\in[a,b]} |F(x) - G(x)|.}
This statistic can be used in the Kolmogorov-Smirnov 2-sample test, which outputs a $p$-value for the probability one would observe
as large a statistic if $F$ and $G$ were sampled from the same
underlying distribution.
We thus define our version of distributional privacy as follows:
\mandc{\fboxsep5pt\fbox{\parbox{0.9\columnwidth}{
\textbf{KS Distributional Privacy ($\pi$-KSDP).}
Mechanism $\mathcal A$ is $\pi-$KS Distributional Private if there exists a simulator mechanism $\mathcal B$ s.t.
for all statistics $\sigma\in\{\sigma_1,\sigma_2,\ldots\}$,
and all documents $x\in~X$,
the KS 2-sample test run on ECDF($\sigma(\mathcal O_{\mathcal A, X,Y},x)$) and ECDF($\sigma(\mathcal O_{\mathcal B, X\setminus \{x\},Y},x)$) returns $p\geq \pi$.
}}}
In words, KSDP means the observables generated by a simulator
that doesn't have a document $x$ cannot be statistically distinguished from the actual algorithm running on the database containing~$x$.
The KS-test's $p$-value is a measure of distance between two distributions which takes number of samples into account.
A high $p$-value means that we cannot reject that the two ECDFs are the result of sampling from the same underlying distribution.
It doesn't mean that we can conclude they come from the same distribution.
Although hypothesis tests with the reverse null/alternate hypothesis have been studied~\cite{wellek2010testing}, computational efficiency and interpretability are significant challenges.

\begin{table}[t]
\centering
\begin{minipage}[t]{0.6\textwidth}
\begin{algorithm}[H]
\caption{Measuring $\pi$-\ksdp.}
\label{alg:posterior_test}
\floatname{algorithm}{Procedure}
\begin{algorithmic}[1]
\renewcommand{\algorithmicrequire}{\textbf{Input:}}
\renewcommand{\algorithmicensure}{\textbf{Output:}}
\Require Mechanism $f$, set of statistics $S$, database $X$, document of interest $x$, number of samples per document $t$.
\Ensure Minimum $p$-value for all statistics.
\State Determine what the \termdef{observables} of $f$ are.
\State Generate database sub-samples $\{X'_i\}_{i=1}^t$,  $X'_i \subseteq X$
\State Evaluate the mechanism $f(X'_i \cup \{x\})$ for all $i$.
\State Generate d.b. sub-samples $\{Y_i\}_{i=1}^t$,  $Y_i\subseteq X\setminus  \{x\}$
\State Evaluate the mechanism $f(Y_i)$ for all $i$.
\State Set $\pi \gets 1$
\For{each $s \in S$}:
\State w/ $\gets$ ECDF(\{$s(f(X_i), x)\}_{i=1}^t$).
\State w/o $\gets$ ECDF($\{s(f(Y_i),x)\}_{i=1}^t)$.
\State $p \gets $ KS 2 sample test on w/ and w/o.
\State $\pi \gets \min(p,\pi)$
\EndFor\\
\Return $\pi$
\end{algorithmic}
\end{algorithm}
\end{minipage}
\end{table}
The definition of $\pi$-KSDP leads to a
method for measuring the privacy of a distributed mechanism
(Algorithm \ref{alg:posterior_test}).
We stress that unlike DP,  $\pi$-KSDP doesn't guarantee
future-proof privacy against new statistics and auxiliary information.
Rather it tests if a given statistic is discriminative enough to break a weaker yet meaningful distributional version of DP.
In practice the most powerful statistic \math{\sigma} is not known. Still,
we may consider a family of plausible statistics and take
\math{\inf_\sigma{\pi\text{-KSDP}(\sigma)}}.
An advantage of Algorithm \ref{alg:posterior_test} is that samples can be generated, stored, and re-used to test many different statistics quickly.
Furthermore, it  is suitable as a defensive algorithm for identifying sensitive documents $x$, which can be excluded from distributed learning if needed.
Effective simulators $\mathcal B$ are to run $\mathcal A$ but either replace $x$ by a random document, or sample $n$ random documents to create an entirely new database.
In our experiments we use such a simulator and Algorithm \ref{alg:posterior_test} to demonstrate that \pd and PD-SVD satisfy $\pi$-KSDP (see Sections~\ref{sec:experimental_results} and~\ref{sec:experimental_results_svd}).

\remove{This means that we need to ensure thatd for each topic~$t$ at each iteration~$i$ the intermediate sums
 $\sum_{m=1}^M W_{:t}^{(m)}R_t^{(m)}$ and $\sum_{m=1}^M ||W_{:t}^{(m)}||_2^2$ are privacy-preserving observables, which we loosely define as


\centerline{\fboxsep10pt\fbox{\parbox{0.925\columnwidth}{
\textbf{Privacy-preserving observable.}
An observable $\mathcal{O}$ is an object that is emitted by party $m$ and 
depends on $X^{(m)}$. We say that $\mathcal O$ is {\it privacy-preserving} if for all rows in the sampling universe, $x\in \mathcal U$, $P(x\in X^{(m)} | \mathcal O) \approx P(x\not\in X^{(m)}| \mathcal O)$.
}}}



The privacy of the observables of our PD-NMF is argued as follows: The privacy of the \ssum protocol ensures that the only observable which might leak information is the sums computed in every round. Because the sum is a symmetric function, this ensures that for any such observable $\mathcal O$ a group of $A$ adversaries  cannot determine which of $M-A$ parties contributed what amount. Hence, except in atypical situations where all or none of $M-A$ have or don't have a particular document, this will ensure that the observable is privacy preserving. Therefore the privacy preserving property reduces to 
%
%
%
%
ensuring that the adversaries cannot determine whether or not an observable is typical. 


One way to obtain this would be  to use a DP mechanism, i.e., we could add noise proportional to the sensitivity of each of our observables w.r.t. the absence or presence of a particular document.
This would indeed protect us against pathological, i.e., atypical, cases of outliers---such as documents that have rare words or users that rate lots of obscure movies--- but in doing so, the sensitivity-induced noise would increase so much due to the outliers that it would negatively impact the learning outcome for `average' users \cite{FastDPMF}.
%
%
One might argue that protecting outliers is the entire point of DP, and we agree. 
 But we contend that pathologically orthogonal outliers can often be identified and are rare by definition. As a result we expect that only a small number of such outliers will be sampled from the universe $U$. For this small number of documents full MPC can be used on along with the solution of PD-NMF for more typical database rows.

Instead of adding noise to our outcomes we demonstrate that the uncertainty of typical scenarios sufficiently conceals the information leaked by the observable sums. The tool best suited for this is the notion of {\it distributional differential privacy}~(DDP)~\cite{bassily2013coupled}. Given a distribution over databases $\Delta$, and a database $X \sim \Delta$, a mechanism $F(X)$ is private if it is $\epsilon, \delta$-indistinguishable from a simulator $S(X_{-i})$ for all $i\in X$; i.e. the simulator with any row of the database removed should produce output that is distributed almost the same as the original mechanism on the entire database. The main issue with this definition is that $\epsilon, \delta$-indistinguishability is a statement that depends on the probability distribution for all subsets of $F$'s output. When $F$ isn't available analytically, experimentally finding $\epsilon, \delta$ becomes analogous to the problem of finding the correct bins for comparing two distributions.}

%% file: experimental_results.tex
\section{Experimental Results}
\label{sec:experimental_results_both}
In this section we experimentally validate  several claims about the accuracy (exactness)  and privacy of our distributed learning algorithms, and compare them with common  technique of applying DP noise in each iteration. We also showcase  our algorithms for classical machine learning problems that can be solved using NMF and SVD.

\subsection{Privacy and Accuracy of PD-NMF}
\label{sec:experimental_results}

\graphicspath{{./}}
{
\setlength{\intextsep}{5pt}%
\setstretch{1}%
\setlength{\belowcaptionskip}{-\baselineskip}\addtolength{\belowcaptionskip}{5pt}

We give empirical evidence to support three claims:
\begin{enumerate}
\item Our private initialization \init yields equivalent results to a non-private \verb+nnsvd+ initialization.
\item The improvement in the learning outcome for cooperative 
distributed learning (vs learning from ones own data)
is significant for both applications we tested: topic modeling (TM),
and recommender systems (RS), especially for small~\math{M}.
\item \pd satisfies $\pi$-KSDP,
while preserving the centralized learning outcome.
\end{enumerate}

{\bf 1. Equivalence of Initialization.}
We compare \init to other distributed and centralized initialization methods for NMF.
The baseline is \verb+best_of_M+ which uses the best set of local
bases $T$ (with minimum total reconstruction error), and \verb+random+.
The competitor algorithms are \verb+nnsvd+ (uses centralized
SVD), and \verb+random 2x+ which starts from random topics and runs
twice as many iterations.
We use the 20NG~\cite{20ng}, Enron~\cite{enron}, and Reuters~\cite{reuters} datasets to compare initialization methods.
We applied stemming and TF-IDF; for the centralized \verb+nnsvd+ we use
global IDF weights, whereas for the distributed initializations
we use local IDF weights.
We varied number of parties $M\in\{3,5,30\}$ as well as the size of each party's local database $f\in\{3\%,10\%,30\%\}$, as a fraction of the underlying corpus. Each party's database doesn't have duplicates, but larger $M$ and $f$ lead to more duplicates in the combined corpus.
We try $k\in\{20, 40, 60\}$ bases and perform 5 trials for each setting.

For the topic modeling application of NMF, me measure the Frobenius reconstruction error on the training corpora, semantic coherence of the topics \cite{roder2015exploring}, as well as perplexity \cite{griffiths2004finding} of the model on held-out test data. We use multiple evaluation measures, because Frobenius reconstruction error doesn't capture all aspects of what it means to find a good topic model. Aggregate averages across all trials and experiment settings are presented in Table \ref{tbl:topic_modeling}, results within $\pm 3\sigma_\mu$ of the best are bolded. 

\begin{table}[h]
\begin{center}
\begin{tabular}{l||r|r|r}
Algorithm&Fro. Error&Coherence&Perplexity\\
\hline\hline best\ of\ M&$8110.7 $&$\mathbf{16.9} $&$44.8$\\
nnsvd&$\mathbf{7903.5}$&$\mathbf{16.9}$&$\mathbf{43.8}$\\
random&$8216.1$&$16.5$&$\mathbf{43.9} $\\
random\ 2x&$\mathbf{7919.8}$&$\mathbf{16.9} $&$\mathbf{43.8} $\\
\init&$\mathbf{7907.9} $&$\mathbf{17.0}$&$\mathbf{43.8}$\\
\end{tabular}
\caption{\init is significantly better than the baselines
and comparable to nnsvd
($\pm3\sigma_\mu$).}
\end{center}
\label{tbl:topic_modeling}
\end{table}

Similar equivalence results hold for the recommender system setting, where the only measure used is RMSE. The main trade-off between the two distributed initialization algorithms \init and \verb+random 2x+ is whether the information emitted during the initialization is more (or less) revealing than the information revealed during additional rounds of PD-NMF.

\emph{Conclusion:}
\init is competitive with state of the art initialization schemes for
\math{NMF}, is efficient, involves fewer rounds of communication and
is private providing \iter is private.

{\bf 2. Improved Learning Outcome.}
PD-NMF must be adapted to a weighted element-wise version
to handle non-observed ratings.
The RRI-NMF
projected gradient updates are only slightly modified,
 as described in Section 6 of~\cite{honmf}.
 The changes can still be accommodated by PD-NMF since these modifications
 fit with communicating via \ssum. To predict ratings using an outcome $T$ of PD-NMF a party $m$ would first fit $T$ to $X^{(m)}$ by
 minimizing $||X^{(m)}-WT||_F^2$ over $W$.
 Then party $m$ predicts rating $(i,j)$ as
 $\min(\max( (WT)_{ij}, \text{rating}_\text{min}), \text{rating}_\text{max})$.
We use the Hyperband algorithm \cite{li2016hyperband} to set
regularization parameters (its key steps only need to run PD-NMF and
\ssum).

\begin{figure*}[t]
\tightcenter
\tabcolsep1pt
\begin{tabular}{p{0.33\textwidth}p{0.33\textwidth}p{0.33\textwidth}}
\includegraphics[width=0.33\columnwidth]{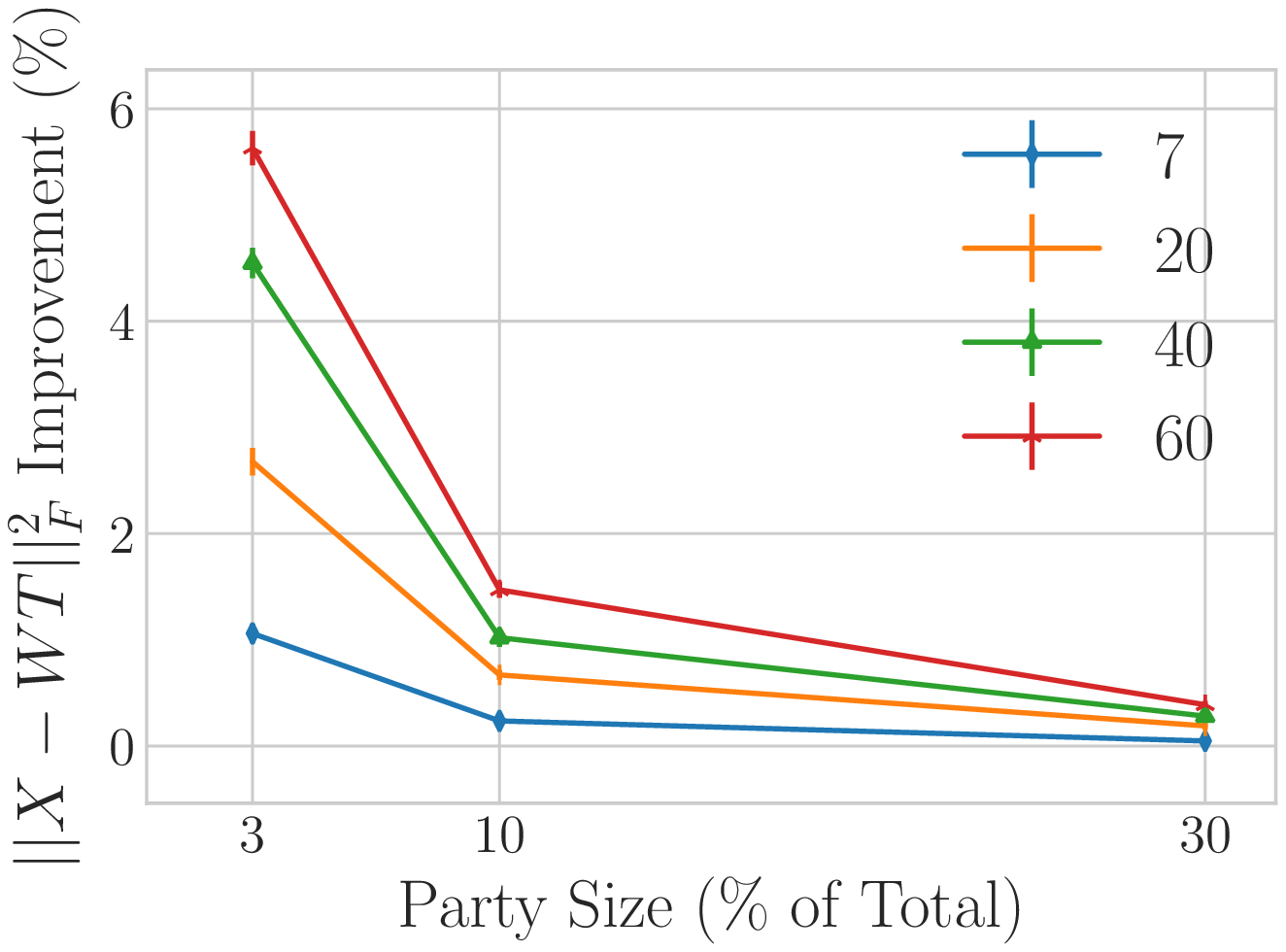}&
\includegraphics[width=0.33\columnwidth]{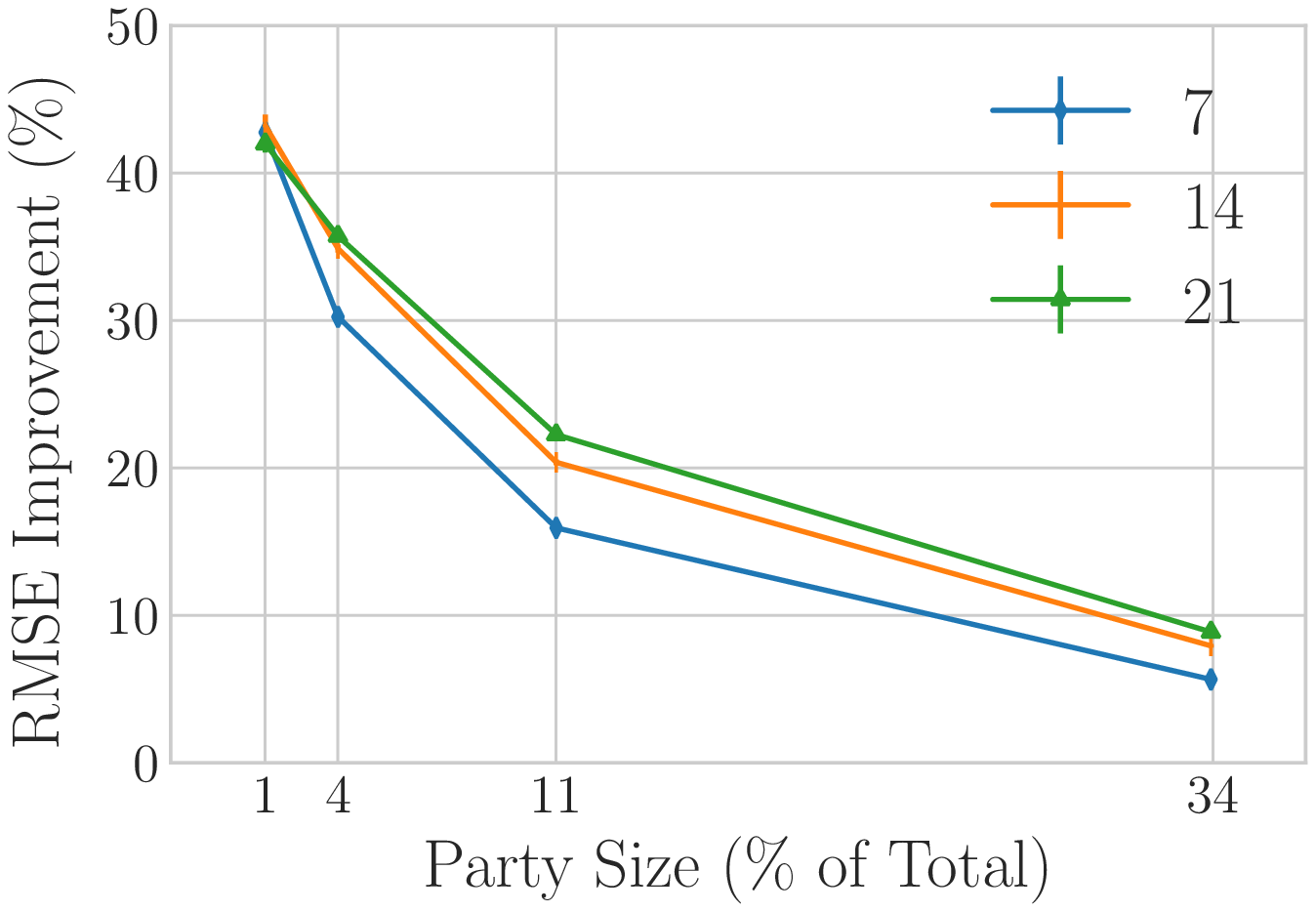}&
\includegraphics[width=0.33\columnwidth]{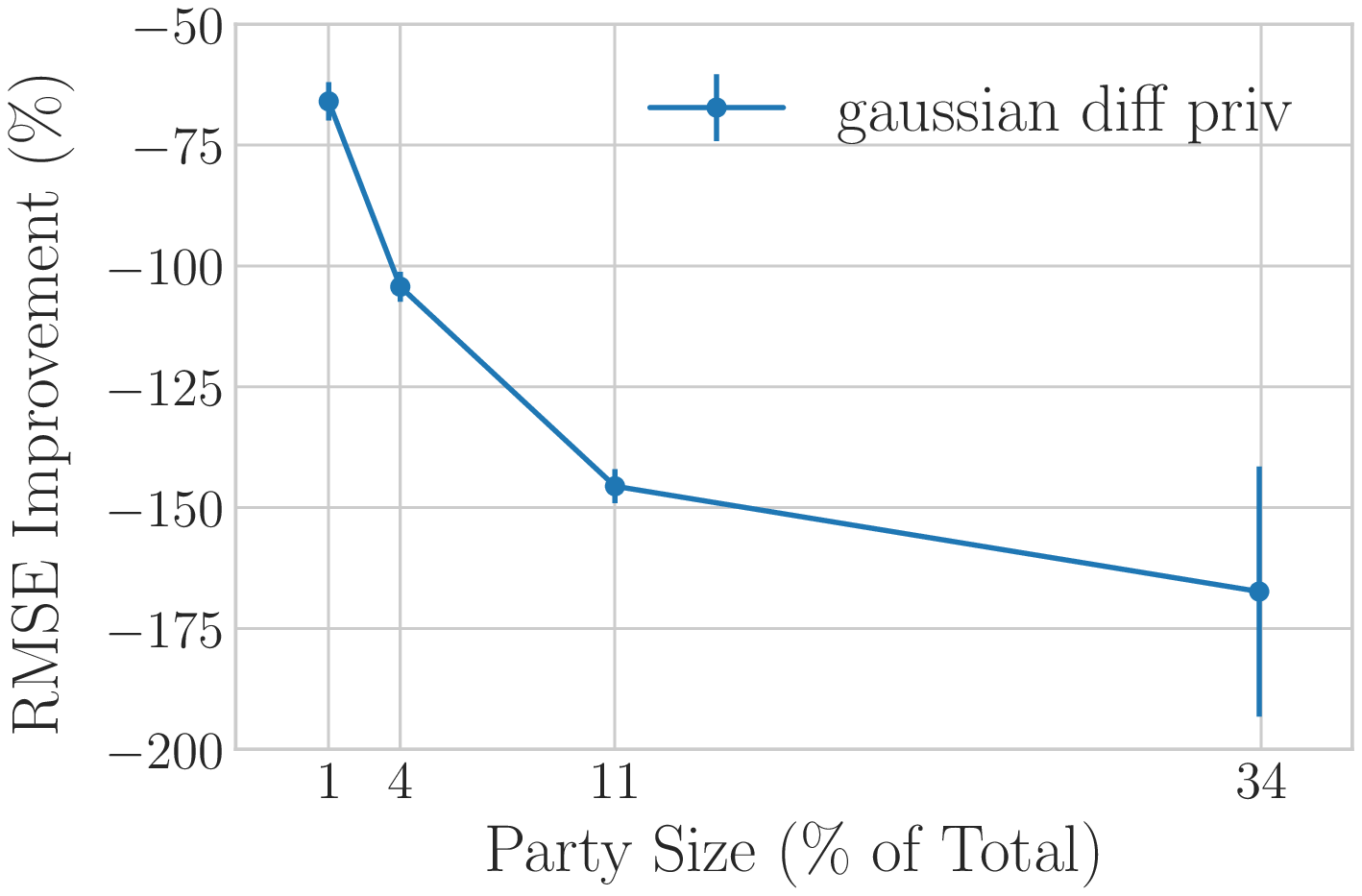}%
\\
(a) PD-NMF (TM)
&
(b) PD-NMF (RS)&
(c) Differential Privacy\\
\end{tabular}
\caption{Percent improvement in the learning outcome.
The different curves are for different values of
\math{k}.
(a) Topic modeling on Enron data.
(b) Recommender systems on ML-1M.
(c) Indicative learning outcome deterioration when $(\epsilon,\delta)$-DP (Gaussian noise for  generous parameters: $\epsilon=0.25$ and  $\delta=0.01$) is used inside the NMF algorithm ($k=7$)  to ensure privacy (see
supplementary material for details).
The cooperative learning
outcome with DP is worse than if the parties do not cooperate and
simply learn on their local data.
This is not surprising due to DP's inherent trade-off of accuracy 
for privacy, which underlines why DP is not suitable for our
setting when the exact learning
outcome is important (such issues with DP have already been raised in
\cite{hua2015differentially,rajkumar2012differentially}).
\label{fig:learning_outcome}
}
\end{figure*}

Next, we can measure \% improvement compared to using local models for both topic modeling and recommender systems. For topic modeling we use the Enron dataset, and measure the percentage decrease in Frobenius reconstruction error using the global topics compared to the best local topics. For recommender systems we use the MovieLens-1M dataset \cite{movielens} and measure the percent decrease in RMSE each party gets for their local users by using the global model instead of their local model. Figure \ref{fig:learning_outcome}  shows that parties of all sizes have something to gain from cooperative distributed learning, but small parties (with the least local data) stand to gain the most.

\emph{Conclusion:}
Significant gains from learning on the combined data are possible, which
is not surprising because more data is better. However, adding noise
to obtain even a moderate level of differential privacy
is detrimental to the learning outcome.

{\bf 2. Privacy of \pd.}
We assume there are $M=2$ parties, and the adversary knows exactly what the victims observables are. In practice, unless colluding with everyone except for the victim, the adversary wouldn't be able to identify a particular party's observables; they would be hidden in the \ssum. The parameters we vary are $f$, the size of each party, between 0.5\% to 10\% of all documents in the given dataset, and the model dimension $k\in\{7,14\}$. We examine this on the Enron dataset for topic modeling and the MovieLens-1M dataset for recommender systems.

To measure privacy of \pd, we use Algorithm \ref{alg:posterior_test}. The simulator runs PD-NMF replacing $X$ with a completely random database $X'$ for which the only requirement is that $x\not\in X'$. The most-discriminative (out of 7 that we tested) document-specific statistic we found is computed as follows: first, find a particular document's coefficients with respect to topics $T$
$\mathbf a \gets \argmin_{\mathbf u \in [0,1]^{1\times k}}||\mathbf x-\mathbf u T||_F^2,$
and then measure the weighted inner product of these coefficients and the observed topic weights (from the denominator in \iter)
$\sigma \equiv \mathbf a\cdot\mathbf w^2.$
We randomly picked 50 documents of interest $x$ (in practice this would have to be done for all documents), and for each we generated 200 samples of PD-NMF running on a random database with $x$, and 200 of it running on a (otherwise~iid) random database without $x$. The former are samples of the original mechanism $\mathcal A$, while the latter correspond to the simulator $\mathcal B(X\setminus\{x\})$. The results are summarized in Table~\ref{fig:KSDP}.

\begin{table}[hbtp]
\centering
\caption{The $p$-value increases as party size increases.}
\ \\ 
\begin{tabular}{l||r|r}
Party Size (\%/100) & RS $p$-value & TM $p$-value\\
\hline\hline 0.005&${0.45} \pm0.16$&${0.029} \pm0.02$\\
0.03&${0.52} \pm0.20$ & ${0.044} \pm0.02$\\
0.1&${0.48} \pm0.18$ & ${0.113} \pm0.07$\\
\end{tabular}

\label{fig:KSDP}
\end{table}

In most practical settings, a \math{p}-value of \math{0.05} is sufficient to
reject the null hypothesis. We see that even at small party sizes, we
may reject the null hypothesis that \math{f} and \math{g} have different
distributions.

\emph{Conclusion:} Privacy is empirically preserved according to
\math{\pi}-KSDP.

%% file: experimental_results_svd.tex
\subsection{Privacy and Accuracy of PD-SVD}
\label{sec:experimental_results_svd}

In this section we give
experimental results to
validate similar claims as {\bf 2} and {\bf 3} above about \pdsvd.
In fact, in addition to proving the improved quality of the learning outcome, we also demonstrate that DP is far less applicable in classical SVD applications. Furthermore, our experimental privacy analysis also highlights the value of hiding $\Sigma$ in the SVD algorithm iterations, a feature that, as discussed in the introduction, distinguishes our approach from existing attempts to private SVD.

\para{Learning Uplift \& Inapplicability of DP}
We show that in two typical applications of the SVD: Principal Component Regression (PCR)~\cite{jolliffe1982note} and Low Rank Approximation (LRA)~\cite{markovsky2011low}. 
By using \pdsvd all participants achieve better models than they would have only using their local data. We also show that differentially private SVD~\cite{hardt2013beyond} has unpredictable learning outcomes, which makes it unsuitable for the $M$-party distributed setting.

In PCR we are given an $n\times d$ feature matrix $X$ and a $n\times 1$ target matrix $y$, and the goal is to predict $y$.
We first transform $X\to \hat X$ by projecting it onto the space spanned by its first $k$ principal components.
Then we find \[
\beta = \arg\min_\beta ||\hat X \beta - y||_2^2.
\]
To find the principal components, the parties use Algorithm~\ref{alg:pdpca} and obtain $V_k$.
They then solve the local PCR problem with the centered $y_0=y\m-\frac{1}{n_m}\sum y_i\m$ \[
\min_\beta \frac{1}{2} ||\left(X\m V_k\right) \beta-y_0||_2^2.
\]
We evaluate PCR by measuring the RMSE each party gets on a held-out $X_\text{test}, y_\text{test}$.
This is a measure of their predictive quality. We report the change in RMSE from using their local $X\m$ to fit $\beta$ compared to using \pdsvd and the combined $X$.

We consider two datasets for PCR. First, the Million Songs dataset~\cite{Bertin-Mahieux2011} wherein the task is to predict the year in which a song was released based on its tonal characteristics.
Second, the Online News Popularity dataset~\cite{fernandes2015proactive} wherein the task is to predict the number of shares a blog post will get based on a variety of heterogeneous features.

For the LRA problem a matrix $X$ is projected onto a rank-\math{k}
subspace using $XV_kV_k^T$. We measure the reconstruction error as \[
||X-XV_kV_k^T||_F,
\]
and compare the reconstruction error each party gets on a held-out $X_\text{test}$ using $V_k$ computed from its local $X\m$ compared to the $V_k$ obtained from the global $X$ using \pdsvd.
We evaluate this on the 20NG~\cite{20ng} and Enron~\cite{enron} text datasets.

The results for PCR and LRA are shown in Figure~\ref{fig:svd_learning_uplift}
In Figure~\ref{fig:svd_learning_uplift} we compare \pdsvd to a particular Differentially Private SVD mechanism~\cite{hardt2013beyond} with parameters $\epsilon=0.1$, $\delta=10^{-5}$.
These parameters are reasonable, even generous, for DP mechanism intended for a database with at least $10^5$ rows.
In three out of four experiment settings, the SVD generated by the DP mechanism yields a worse learning outcome than each party could have attained by simply using their own local data.
However for the Online News Popularity dataset, differentially private SVD yields an uplift for all but the largest party sizes.
This highlights the issue that despite the fact that we use the same DP mechanism across all settings, the effect of differential privacy needs to be analyzed on a case-by-case basis.
Because of this analysis challenge, and since on 3/4 settings the parties got no learning benefit, and lost some privacy, we believe DP is an unattractive method for build privacy-preserving $M$-party distributed mechanisms.

\emph{Conclusion:}
The overall trend for both applications, and all datasets, is that smaller parties, which have less local data, have the most to gain from participating in distributed learning. Differential privacy has worse learning outcome, and can even
be worse than the local learning outcome.

\newcommand{\putlabels}[4]{
  \begin{tikzpicture}
    \node[scale=#2,inner sep=0pt](P)at(0,0){\includegraphics*{#1}};
    \node[scale=0.8,anchor=north,inner sep=0pt](xlabel)at($(P.south)+(0,0.05)$){#3};
    \node[scale=0.8,anchor=south,inner sep=0pt,rotate=90](ylabel)at($(P.west)+(0.1,0)$){#4};
  \end{tikzpicture}
}

\begin{figure}[tbhp]
\caption{Average improvement in learning outcome. The $y$-axis shows the \%
improvement from using global $X$ compared to local $X$ for PD-SVD and Differentially Private SVD for different party sizes. The \math{x}-axis shows the party
size as a percent of all the rows in the datas.
Naturally, the value add of participating in the distributed protocol
decreases as a party's local data increases. 
\label{fig:svd_learning_uplift}}
\begin{center}
\tabcolsep5pt
\begin{tabular}{@{\hspace*{0pt}}llll}
\multicolumn{2}{c}{\textsc{Principal Component Regression}}&\multicolumn{2}{c}{\textsc{Low Rank Approximation}}\\\hline
\textsc{Million Songs} & \textsc{Online News}&\textsc{20NG} & \textsc{Enron}\\
\putlabels{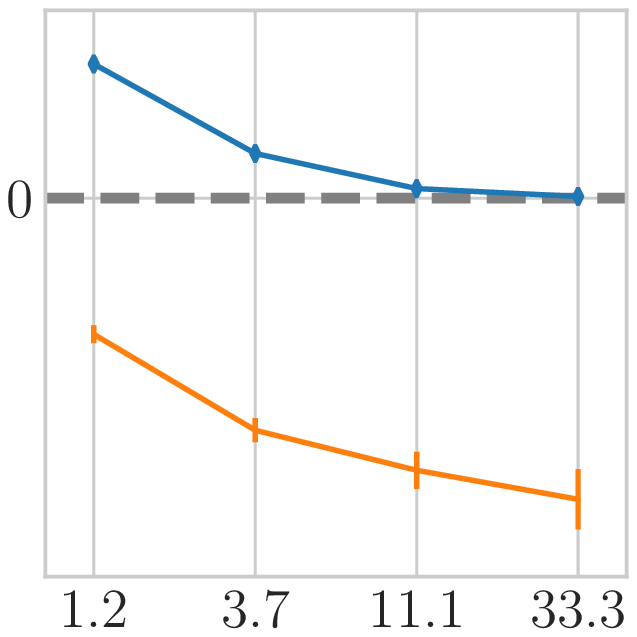}{0.5}{party size (\%)}{Learning-uplift (\%)}
&
\putlabels{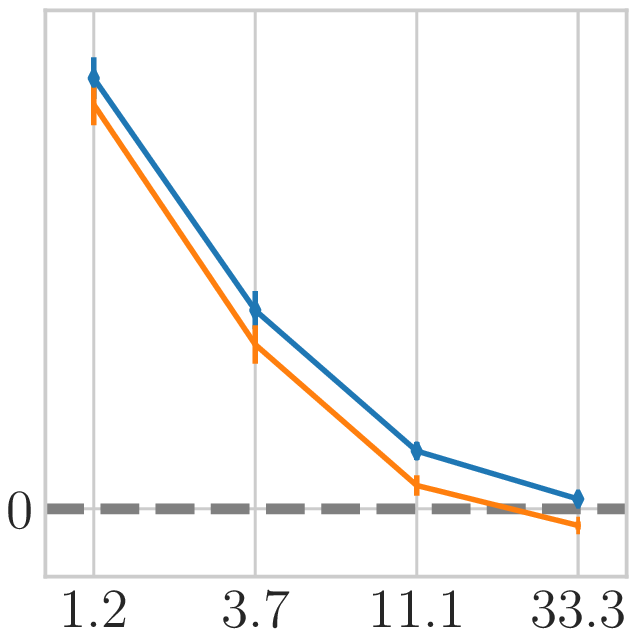}{0.5}{party size (\%)}{Learning-uplift (\%)}
&
\putlabels{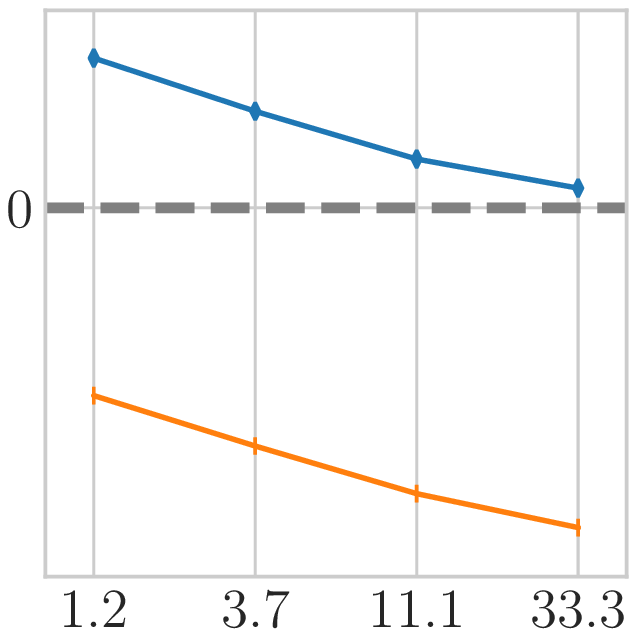}{0.5}{party size (\%)}{Learning-uplift (\%)}
&
\putlabels{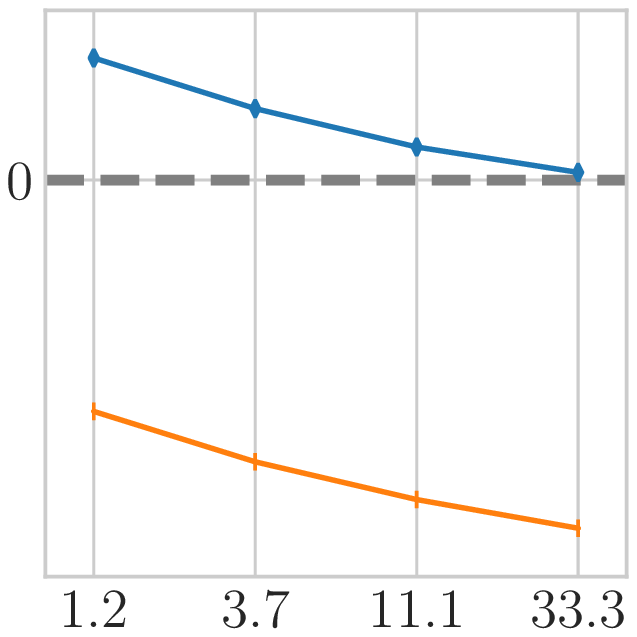}{0.5}{party size (\%)}{Learning-uplift (\%)}
\\
\multicolumn{4}{c}{\includegraphics[ height=1.5em]{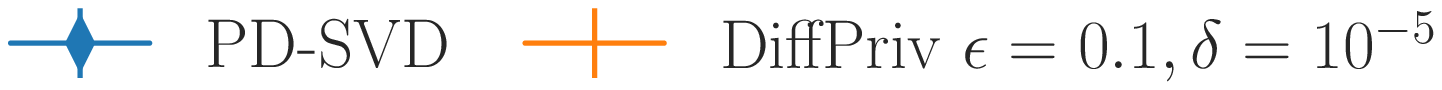}}
\end{tabular}
\end{center}
\end{figure}

\para{Measured Privacy \& Value in Hiding $\Sigma$}
We show that \pdsvd has high $p$-values as measured by $\pi$-\ksdp. We also show that \pdsvd is more private than $\Sigma$-revealing SVD algorithms that use secure multiparty computation but expose $\Sigma$ in addition to $V_k$.~\cite{SEC:SEC1501,DBLP:journals/corr/ChenLZ17a}.

To measure the privacy of \pdsvd we use Algorithm~\ref{alg:posterior_test}.
We consider 200 documents $x$ (in practice a party would run this for all their documents, or the ones they are worried about), and generate 50 sample databases that have $x$ and 50 (otherwise independent) databases that don't have $x$.
We then run \pdsvd as well as $\Sigma$-revealing SVD on each of the sample databases, and collect all the observables that could be inferred from the communication transcript.
Although \pdsvd is an iterative algorithm, we only study the observables at the end of all iterations, the final $V_k$.
We make this simplification for several reasons: (1) a polynomially-bound adversary can not easily work with time-series of observables due to the curse of dimensionality (2) these observables are highly auto-correlated, meaning that observing $V_k$ over $\tau$ iterations doesn't give one much more information than observing $V_k$ once (3) the observables over multiple iterations can be simulated by working from $X^TX$, so that the privacy of statistics computed from $X^TX$ is a lower bound for statistics computed from multiple iterations' worth of $V_k$s. Since $X^TX=V\Sigma^2V^T$, this lower bound comes from evaluating the $\Sigma$-revealing SVD.

\begin{figure}[tbhp]
\centering
\tabcolsep0pt
\begin{tabular}{c@{\hspace*{0.15\columnwidth}}c}
\putlabels{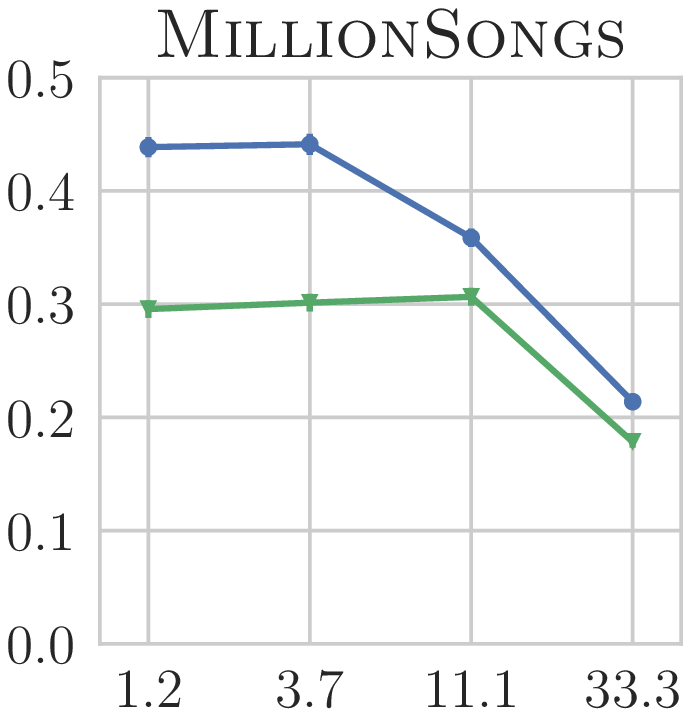}{0.65}{party size (\%)}{KSDP \math{p}-value}
&
\putlabels{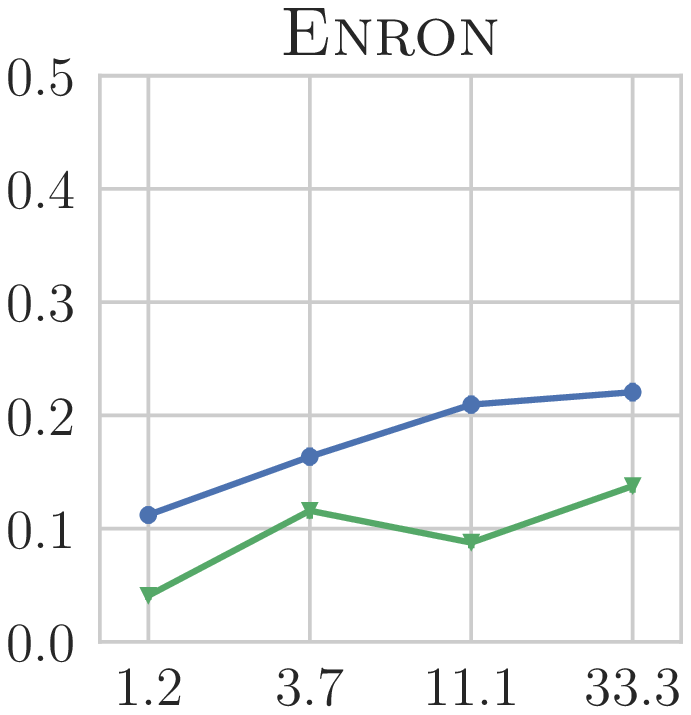}{0.65}{party size (\%)}{KSDP \math{p}-value}
\\
\multicolumn{2}{c}{\includegraphics[width=0.42\textwidth]{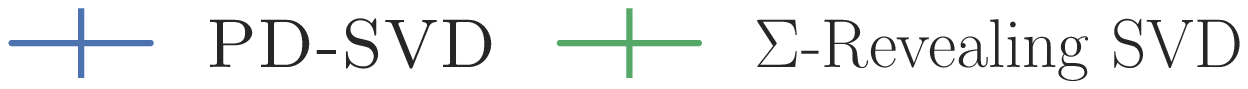}}
\end{tabular}
\caption{The \math{y} axis is the minimum $p$-value as measured by
\ksdp, versus the party's size on the $x$-axis.
Large \math{p}-values are observed, which means privacy is preserved. Further,
privacy significantly drops if \math{\Sigma} is revealed, hence
\nssum is essential.
\label{fig:ksdp_svd}}
\end{figure}

We attempt to simulate a adversary who is clever about how they build
statistics.
First, they observe $V_k$ as an output from the protocol.
Then for \pdsvd they attempt to estimate the global $\Sigma_k^2$ from the knowledge of how many documents the victim has $n_v$.
They know their own $\Sigma_k^{(A)}$, and estimate $\hat \Sigma_k^2 = \frac{n_v+n_a}{n_a} (\Sigma_k^{(A)})^2.$ From this the adversary estimates $S=X^TX$ as $\hat S = V_k \hat \Sigma_k^2 V_k^T$.
For $\Sigma$-revealing SVD the adversary simply uses the revealed $\Sigma$ rather than this estimate.
Since the adversary is building a statistic for a particular document $x$, they can consider the relationship between $\hat S$ and the outer-product $xx^T$.
We derived one family of statistics from the weighted correlation matrix $W=|\hat S \circ{} xx^T|$, where $\circ$ and $|\cdot|$ represent elementwise product and absolute value, respectively.
To reduce $W$ to a scalar we consider taking the max entry, as well as sum, and sum of squares.
We also consider a similar family of statistics originating from $D=\left|\frac{\hat S}{||\hat S||_F} - \frac{xx^T}{||x||}\right|$.
In Figure~\ref{fig:ksdp_svd} we present the mean, over documents $x$, minimum $p$-value resulting from both families of statistics.

For the LRA task on Enron, we see that small parties are at highest risk for privacy breach.
However, since in \pdsvd observables are communicated through \ssum and \nssum mechanisms a coalition of $A$ colluding parties would only be able to observe the sum of the remaining $M-A$ non-colluding parties' observables.
As a result, a party's `effective party size' would be the sum of the non-colluding parties' sizes.
For both PCR on Million Songs and LRA on Enron, we observe that \pdsvd has a significantly higher $p$-value than the $\Sigma$-revealing SVDs, except for the 33.3\% party size on Enron.
This confirms the intuitive idea that $\Sigma$ contains informative information.
However, since even the $\Sigma$-revealing SVD has $p$-values that are large, so running \pdsvd with \ssum may be acceptable for many applications.

\emph{Conclusion:} Privacy is empirically preserved and concealing
\math{\Sigma} using \nssum preserves considerably more privacy.

%% file: conclusion.tex
\section{Conclusion}
\label{sec:conclusion}

We are proposing a new framework for distributed
machine learning when the learning outcome must be preserved, which strikes a compromise between efficiency and
privacy. On one end of the spectrum
sits general MPC which is impractical but guarantees
almost no information-leak.
On the other end is complete data sharing which also preserves
learning
outcome but all privacy is lost.
Our approach sits in the middle:
 modify the learning algorithm so that efficient
lightweight crypto can be used to control the information leaked
through only those few shared aggregate quantities. Then, {\it measure}
the information that is leaked.

Overall there is an interesting trade-off between privacy and learning outcome in distributed learning.
Here we emphasized the learning outcome because
there are significant performance gains from learning
on all the data as opposed to learning locally. In particular 
smaller parties gain more in learning outcome, but
also risk losing more privacy. 
We give a framework for mitigating this and apply it in two significant learning algorithms, namely NMF and SVD. In our protocols, all communication is done via lightweight MPC modules 
(\ssum and \nssum) hence the effective party size
a coalition of adversaries sees is the sum of the
remaining parties. 

As with MPC, our algorithms preserve the learning outcome.
We ensure 
that 
information is only leaked through the {aggregate} (normalized) outputs
shared by \ssum and \nssum,  and from the learning output---no other information is exchanged or leaked. We introduce the notion of 
KS-Distributional Privacy to  {experimentally estimate} the leakage from announcing these aggregate-outputs/sums.
Note that
KS-Distributional Privacy can also be used to estimate, in practice, the
privacy of the output of an MPC and hence is of independent interest.

We believe that, in practice, economic incentives make either privacy or learning outcome the top priority for a particular group of parties.
As a result, tools that put the learning outcome first, but provide a way to measure privacy against a pre-defined or iteratively updated set of post-processing functions are useful.
It's reasonable to expect organizations to be able to define post-processing methods for which they want privacy of their documents to hold.
Legislation such as the GDPR describes post-processing methods, against which data must be secure.
Meanwhile, information security guidelines such as ISO-27001 emphasize the need to iteratively evaluate security needs, which can include post-processing methods.

Future research includes other empirical methods for estimating privacy, automated ways to build discriminative statistics, and a complete optimized library of elementary MPC protocols for machine learning.

%% file: appendix.tex
\appendix

\section{MPC Versus DP in the context of Machine Learning}
\label{appendix:MPCvsDP}

We give a comparison of differential privacy (DP)
with multiparty computation (MPC).
\label{table:DP-MPC}
\begin{center}
\begin{tabular}{c|p{0.4\linewidth}p{0.4\linewidth}}
&Differential Privacy& Multiparty Computation\\\hline
Who&Data owner and public& \math{M} mutually distrustful parties.
\\
Task&Publicly answer query on data.
&
Compute a function $f$ on the joint data.
\\
How&Output is noised to protect records. 
&
Distributed cryptography to compute $f$. 
\\
Computation&Private, light.&Distributed, peer-to-peer, heavy.
\\
Privacy& Public cannot infer any database record after observing
answer to the querry.
&
Parties cannot infer anything {\it beyond} what
is inferred from the exact output.
\end{tabular}
\end{center}

\remove{
The following table summarizes  our work in relation to prior work.
\begin{center}
\begin{tabular}{p{0.2\columnwidth}p{0.25\columnwidth}p{0.35\columnwidth}}
\textsc{Data ownership}   & \multicolumn{2}{p{0.7\columnwidth}}{\textsc{Privacy against an adversary whose post-processing power is}} \\

		& \multicolumn{1}{c}{\textit{Limited}} & \multicolumn{1}{c}{\textit{Arbitrary}} \\ \cline{2-3} 
 \multicolumn{1}{r|}{
 \textit{One Party}
 }  

& \multicolumn{1}{p{0.25\columnwidth}|}{Non-private NMF E.g. \cite{gemulla2011large,du2014maxios}
} & \multicolumn{1}{p{0.35\columnwidth}|}{Differential Privacy}             \\ \cline{2-3}

 \multicolumn{1}{r|}{\textit{$M$ Parties}} & 
 \multicolumn{1}{p{0.25\columnwidth}|}{\cellcolor[HTML]{7CEB98} \textbf{PD-NMF} and Secure MPC, e.g. \cite{Kim2016,nikolaenko2013privacy}.
 } & \multicolumn{1}{p{0.35\columnwidth}|}{Multi-party Differential Privacy}             \\ \cline{2-3} 
\end{tabular}
\end{center}
}

\section{Proofs Deferred to the Appendix}
\label{app:omitted_proofs}

\subsection{Proof of Theorem~\ref{thm:pdnmfiter}}
\begin{proof}
Algorithm \ref{alg:distributed_rri_nmf} was derived by breaking up a matrix-vector product
into a sum of sub-matrix vector products, and this forms the basis for this result. Note that $T$ and the regularization constants are known to all parties for all iterations.

First, for the update of $W$ in
 \r{eq:nmf_update}, let $I_m$ be the rows of $X$ that belong to party $m$.
Then in centralized \textsc{ \textsc{RRI-NMF}} the update for \math{W_{I_mt}^{(m)}} depends on \math{(R_t[I_m,:]T_{t:}^T)^{(m)}} and globally known information. Since $R_t[I_m,:]=R_t^{(m)}$, each party can compute this locally,
which gives step 4 in the algorithm.

The update to the topics $T$ requires
\math{\norm{W_{:t}}_2^2} and \math{W_{:t}^TR_t} which are given by sums,
\mandc{
\norm{W_{:t}}_2^2=\sum_{m=1}^M\norm{W_{:t}^{(m)}}_2^2
\,\text{;}\,
W_{:t}^TR_t=\sum_{m=1}^MW_{:t}^{(m)T}R_t^{(m)}.}
(\math{R_t^{(m)}} is party \math{m}'s residual matrix.)
Since these sums are computed using \ssum in steps 5, 6 we know that no additional information is leaked to the network, and that the output is correct. Step 7 combines these sums in the same way as the update in Equation \ref{eq:nmf_update}.
\end{proof}

\subsection{Proof of Theorem~\ref{theorem:upper}}

\begin{proof}The upper bound result follows from setting $V$ (step 4 of \init) by analyzing $\frac{\partial||X-WT||_F^2}{\partial T_{t:}}$. 
This is a simple application of the Taylor expansion along with the assumption that local NMF models (step 2 of \init) have converged; see Appendix A for details.

For privacy, suppose \iter is private on input $\{X^{(m)}\}_{m=1}^M$ initialized with $T_0$
(argued in Sections \ref{sec:privacy_analysis}, and \ref{sec:experimental_results}).
The only step of \init where something non-random is communicated to the other parties is step 9.
However, since $\hat T^{(m)}$ is a deterministic function of $X^{(m)}$, step 9 can be simulated using \iter~($\{X^{(m)}\}_{m=1}^M, T_0$), which is private by assumption.
\end{proof}

\subsection{Correctness of Initialization Algorithm}
\label{app:weighted_merge}

\begin{claim}
Algorithm 2 minimizes an upper bound on the reconstruction error $||X-WT||_F^2$ by using $T_{in}$ as a proxy for $X$.
\end{claim}
\begin{proof}
The goal is to bound the magnitude of the residual (denoted as $d_t$) left by reconstructing the $t$-th topic $T_t$ as $\hat T$ minus an error vector $\Delta_t$. In our case $$d_t(T_{t:})=||R_t-W_{:t}T_{t:}||_F^2.$$ 

The quantity we are interested in is $d_t(\hat T_t) = d_t(T_t+\Delta_t).$

We can expand the RHS using Taylor's Theorem,

\begin{equation}
d_t(T_t+\Delta_t) = d_t(T_t) + \left(\frac{\partial d_t}{\partial T_t}\right)^T  \Delta_t+ 0.5  \Delta_t^T \left(\frac{\partial^2 d_t}{\partial T_t^2}\right) \Delta_t. \label{eq:dt}
\end{equation}

Note that $\frac{\partial d_t}{\partial T_t} = ||W_t||^2T_t-W_t^TR_t$, and consequently $\frac{\partial^2 d_t}{\partial T_t^2} = I_d||W_t||^2$.

Then we can rewrite Equation \ref{eq:dt} as 

\begin{equation}
d_t(T_t+\Delta_t) = d_t(T_t)
+ ||W_t||^2T_t^T\Delta_t-R_t^TW_t\Delta_t
+ 0.5  ||\Delta_t||^2||W_t||^2.
\label{eq:dt_expanded}
\end{equation}

The next step is the only approximate one, where we assume that all local models have converged, and therefore $\frac{\partial d_t}{\partial T_t}=0$. As a result we get $T_t=W_t^T*R_t/||W_t||^2$. Plugging this back into \ref{eq:dt_expanded}:

\begin{align*}d_t(T_t+\Delta_t) &= d_t(T_t)
+ ||W_t||^2T_t^T\Delta_t-||W_t||^2T_t^T\Delta_t
+ 0.5 ||\Delta_t||^2||W_t||^2\\
&= d_t(T_t) + 0.5 ||\Delta_t||^2||W_t||^2
\end{align*}

This means that a $\ell_2$-reconstruction error of $||\Delta_t||^2$, for topic $t$ will incur a cost of at most $||W_t||^2=\sum_{i=1}^n W_{it}^2$ in terms of error on the original $R_t$ and consequently original $X$. In the last step of Algorithm 2 when solving $\frac{1}{2}|| \text{diag}(\mathbf v)(T_{in}-WT)||_F^2$, it is precisely the $||\Delta_t||^2$s that are minimized.\end{proof}

\subsection{Proof of Theorem~\ref{thm:svd}}

\begin{proof} In order to derive a private version of Algorithm~\ref{alg:non_private_power_iteration} we need to perform the \termdef{initialization} (step 2), \termdef{iteration} (step 3) in a private and distributed way.

For initialization, each party $m$ samples
\mandc{
\hat V^{(m)}\sim \mathcal N\left(0,\frac{1}{\sqrt M d}\right)^{d\times k}
\text{, so that }
\sum_{m=1}^M \hat V^{(m)} \sim M\mathcal N\left(0, \frac{1}{\sqrt M d}\right)^{d\times k} = \mathcal N\left(0, \frac{1}{d}\right)^{d\times k}.
}

For iteration first note that step 3 actually reveals $\Sigma_k$, i.e. $\hat\sigma_i = ||\hat V_{:i}||_2$.
However these sums are not necessary for the algorithm, since they are destroyed by orthonormalization (step 4).
As a result we can hide them using \nssum.
To do so, note that $S$ can be broken down into $M$ sums,
\mandc{
\begin{aligned}
S &= X^TX = \sum_{i=1}^n X_{i:}^TX_{i:} = \sum_{m=1}^M \sum_{j=1}^{n_m} \left(X_{j:}^{(m)}\right)^T X_{j:}^{(m)}\\
    &= \sum_{m=1}^M \left(X^{(m)}\right)^TX^{(m)}.
 \end{aligned}
}
Then, the $i$th column of step 3 can be calculated as
\mandc{
\nssum\left(\left\{\left(X^{(m)}\right)^T X^{(m)}\hat V_{:i}\right\}_{m=1}^M\right).
}
We simply repeat this (or execute it in parallel) for all $k$ columns.
The orthonormalization should be performed using the same algorithm by each party, e.g. the modified Gramm-Schmidt~\cite{trefethen1997numerical}.
\end{proof}

\section{Details of \ssum}
\label{app:smsp}
\paragraph{An Intuitive Example} For convenience we describe the sum protocol in an example with m=3 players. The rows are the sharings of each party $p_m$'s input (the numbers horizontally sum up to the input). The columns are the shares that this column party knows (the $i$th column consists of the shares of party $p_i$ for each of the three inputs $x_1, x_2, x_3$.) The vertical (column's) sum is the share of the sum for this column's party. All the summations are modulo $L$ (which is omitted for presentational simplicity). 

\begin{tabular}{lc |   lll | l}
 &                & \multicolumn{3}{c}{Known by party}   \\\hline
 & Known by party & 1              & 2           & 3           & Row $\sum$s to:  \\\hline
 & 1              & $x_1^{(1)}$    & $x_1^{(2)}$ & $x_1^{(3)}$ & $x_1$ \\
 & 2              & $x_2^{(1)}$     & $x_2^{(2)}$ & $x_2^{(3)}$ &  $x_2$\\
 & 3              & $x_3^{(1)}$    & $x_3^{(2)}$ & $x_3^{(3)}$ &  $x_3$\\\hline
 & All               &  $X^{(1)}=\sum_{m=1}^M x_m^{(1)}$              & $X^{(2)}=\sum_{m=1}^M x_m^{(2)}$            &  $X^{(3)}=\sum_{m=1}^M x_m^{(3)}$           & $X=\sum_{m=1}^M x_m$
\end{tabular}

Correctness of the protocol follows from the fact that the sum of the sum of rows (which is the sum we want to compute) equals the sum of the sum of columns (which is how we compute it). 

\paragraph{Theory} Consider one input element \math{x^{(m)}}
from each party $p_m$. The task is to securely compute their sum
(we can trivially extend to sums of vectors or matrices by invoking
the same protocol in parallel for each component).
\ssum relies on a standard cryptographic primitive, known as {\it M-out-of-M (additive) secret sharing} which allows any party acting as the {\it dealer}, wlog party $p_1$, to share his input $x^{(1)}$ among all parties so that no coalition of the other $M-1$ parties learns any information about $x^{(1)}$.

Let $L-1$ be an upper bound on the size of numbers involved in the sum we wish to securely compute.\footnote{In particular, this is the case in
fixed precision arithmetic.}
To share $x^{(1)}$, the dealer $p_1$ chooses $M$ random numbers $x_{1}^{(1)}, \ldots, x_{M}^{(1)}\in\{0,\ldots,L-1\}$  so that  $\textstyle\sum_{m=1}^{M} x_m^{(1)}=x^{(1)} \pmod{L}$. \footnote{For example choose $x_{2}^{(1)},\ldots, x_{M}^{(1)}\sim
U[0,\ldots,L-1]$ and set
$x_{1}^{(1)}=x^{(1)}-\sum_{m=2}^{M} x_m^{(1)} \pmod{L}$.}
The dealer \math{p_1} now sends $x_{m}^{(1)}$ to
each party $p_m$ ($x_{m}^{(1)}$ is \math{p_m}'s {\it share}  of $x^{(1)}$).
One can show that any coalition  of parties that does not include
the dealer gets no information on $x^{(1)}$.
Each party \math{p_m} acts as dealer and distributes a share of \math{x^{(m)}}
to every party.
%
Party \math{pm} can  now compute his share \math{S^{(m)}}
of the {\it sum}
by locally adding his shares of each $x^{(j)}$,
\math{S^{(m)}=\sum_{j=1}^Mx^{(j)}_m} .
Since this operation is local, it leaks no information about any value.
To announce the sum, the parties need only announce their sum-shares $S^{(m)}$
and then everyone computes the sum as $\sum_{m=1}^{(M)}S^{(m)} \pmod{L}$. 
Note that instead of \ssum, one could use additive homomorphic encryption, e.g., \cite{Paillier:1999:PCB:1756123.1756146}, but this would incur higher computation and communication.   

\paragraph{Complexity} Each of the additions requires one
to communicate  twice as many elements as the terms in the sum---one for each party sharing his term and one for announcing his share of the sum. However, when several secure sums are performed in sequence (as is the case in our algorithm) we can use a pseudorandom generator (PRG) to reduce the communication to exactly  as many  elements as the terms in the sum per invocation of \ssum, plus a seed for the PRG communicated only once in the beginning of the protocol

\paragraph{Implementation-specific Improvements} We next show an optimization in computing the sum by employing a pseudo-random function (or, equivalently, a pseudo-random generator). A pseudo-random function $F_k(y)$  is seeded by a short seed (key) $k$ and  for any distinct sequence of inputs, it generates a sequence of outputs that is indistinguishable from a random sequence (this, effectively, means that is can be used to replace a random sequence in any application). I.e., If the key $k$ is random then  $F_k(1), F_k(2), \ldots$ is a randomly looking sequence.

In order to improve the communication complexity, we use the following trick in the sum protocol: 
\begin{itemize}
\item Each $p_m$, sends (only once, at the beginning of the protocol) to every $p_i$ a random seed $k_{mi}$ to be used for the PRF $F$. 
\item For each $j$th invocation of the secure sum protocol, instead of every $p_m$ sending a fresh random share to each $p_i$, they simply both compute this share locally as $F_{k_{mi}}(j)$. Since the sequence $F_{k_{mi}}(1),F_{k_{mi}}(2),\ldots$ is indistinguishable from a random sequence,  the result is indistinguishable from (and there for as secure as for any polynomial adversary) the original setting where $p_m$ chooses these values randomly and communicates them to $p_i$ in each round. 
\end{itemize}

Note that in modern processors, we can use AES-NI which computes AES encryptions in hardware to instantiate such a PRF with minimal overhead. (The assumption that AES behaves as a PRF is a standard cryptographic assumptions). 

As an example of the gain we get by using the above optimization we mention the following: Assuming a fully-connected topology, for each topic and each iteration each party would have to send and receive $(M-1)\times(d+1)$ floats, for a total communication of $2(M-1)\times(d+1)\times 4$ bytes. For example if $d=1e4$, $M=5$ there would be $0.32$MB per topic per iteration. Even though this is a modest amount, our distributed algorithm is communication bottlenecked because RAM and the CPU operate at a much higher bandwidth than the network.

%% file: supp_svd.tex

\section{Extension to PCA}
\label{app:pca}
The \termdef{principal components} of a matrix $X$ are the eigenvectors of its covariance matrix.
We define the \termdef{expectation} of a matrix $X$ as the mean of its rows
\[
E[X] = \frac{1}{n} \sum_{i=1}^n X_{i:}.
\]
Let $\tilde X=X-E[X]$ be the \termdef{centered} version of $X$.
The \termdef{covariance matrix} of a matrix $X$ is defined as \[
C[X] = E[\tilde X^T\tilde X] = \frac{1}{n}\sum_{i=1}^n \tilde X_{i:}^T \tilde X_{i;} = \frac{1}{n}\tilde X^T\tilde X.
\]
The top-k principal components of of $X$ are the top $k$ right singular vectors of $X$, and these are invariant to scaling of $C[X]$, i.e.
\[
\text{ if } C[X]\approx V_k \Sigma_k^2 V_k \text{ then } sC[X] \approx V_k \hat \Sigma_k^2 V_k.
\]

We rewrite $E[X]$ in terms of $M$ parties \begin{align*}
E[X] &= \frac{1}{n}\sum_{i=1}^n X_{i:} \\
	&= \frac{1}{n}\sum_{m=1}^M \sum_{j=1}^{n_m} X_{j:}\m\\
	&= \frac{1}{n}\sum_{m=1}^M n_m E[X\m].
\end{align*}
Thus if we know $n=\sum_{m=1}^M n_m$, each party can subtract an appropriate multiple of $E[X\m]$ from its $X\m$,
and then use \pdsvd to compute the principal components of $X$.
We present the complete algorithm as Algorithm~\ref{alg:pdpca}.
Note that step 4 computes the principal components of $n^2C[X]$, which are the same as the principal components of $C[X]$.

\begin{algorithm}[tbp]
\begin{algorithmic}[1]
\caption{Private-Distributed PCA, party $m$'s view.}
\label{alg:pdpca}
\floatname{algorithm}{Procedure}
\Require $X^{(m)}$: $n_m\times d$ data matrix, access to \ssum, acces to \pdsvd.
\Ensure $\hat V_k$: estimates of the top $k$ principal components of $X$.
\State $n \gets$ \ssum($n_m$)
\State $\mbf\mu\m \gets \frac{n_m}{n} \sum_{i=1}^{n_m} X_{i:}\m$
\State $\hat X\m \gets X\m - \mbf\mu\m$
\State Invoke \pdsvd($\hat X\m$).
\end{algorithmic}
\end{algorithm}

\section{Normalized Secure Sum and its Implementation via SPDZ}
\label{app:spdz}

In this section we discuss the Normalized Secure Sum and give details on its implementation and security analysis (and the related challenges.)\ \\ \medskip 
\centerline{
\fbox{
\parbox{0.935\textwidth}{
\textbf{\nssum.} Each party inputs a vector $\mbf x_m$. Let $\mbf s=\sum_{m=1}^M \mbf x_m$ be the sum of the inputs.
The output to each party is the normalized sum
$$\nssum(\{\mbf x_m\}_{m=1}^M)=\frac{\mbf s}{||\mbf s||_2}.$$
{\tabcolsep3pt
There exist implementations of \nssum which are:\\
\begin{tabular}{lp{0.8\textwidth}}
\textit{Correct}&The output is the same up to a specified precision as if it was computed centrally.\\
\textit{Secure}&Each party only learns the output and information deducible from the output.\\
\textit{Efficient}&In the online phase, communication and computation cost is $O(Md)$, the same asymptotic cost as the insecure variant. 
\end{tabular}
}
}}}\medskip

\para{The main challenges}
While SPDZ enables implementation of secure protocols, it typically involves non-trivial tuning to fit the problem at hand.
The first issue is that securely computing functions of integers is much more efficient than floats.
In SPDZ, floating point operations incur a considerable slowdown.
The main reason is that all operations in SPDZ are field operations on elements of a finite field, and in the implementation of floating point arithmetic, the privacy of some of the parameters---e.g., number of shifts in a register---needs to be protected to ensure the privacy of the overall computation. We overcome this issue by using fixed point numbers. We fix precision at $f$-bits and a number $x \in \mathbb{F}_p$ has value $x \cdot 2^{-f}$.
Next, using loops is challenging. When loop or branching constructs are used, any variable that needs to be written to, must be in a well defined memory location in the `distributed virtual machine'. This means that ordinary assignments to variables do not work.
Finally, while SPDZ provides support for common arithmetic operations, we needed to implement `square root' ourselves.

\para{Implementation}
\nssum is implemented by expressing Algorithm~\ref{alg:nssum} in the SPDZ language, as follows.
\underline{Input/Output}: Each party $m\in\{1,\ldots,M\}$ computes an additive sharing of its input $x_m$---i.e., random values/shares $x_m[1], \ldots, x_m[M]$ that sum up to $x_m$---
    and sends to each party $p_j$ his share $x_m[j]$. Outputs $\mathbf{\hat s}$ are revealed to all parties. (Input and Output are converted from/to floats to/from a format appropriate for the framework).
\underline{Loops at L \#1, \#2, \#6, \#10} are implemented using the {\tt @for\_range} primitive in SPDZ.
\underline{Variables inside loops}: are globally defined memory locations, e.g. the {\tt MemValue} type in SPDZ. 
\underline{Square root at L \#9}: To compute square root of a number $S$ where $S$ can be expressed as $S = a \times 10^{2n}; 1 \leq a < 100 \text{ and } n \text{ is an integer}$. We first approximate it with  
    \[
    {\sqrt{S}} \approx \begin{cases} 
        2 \cdot 10^n & \mbox{if } a < 10 \\ 
        6 \cdot 10^n & \mbox{if } a \geq 10 
        \end{cases}
    \] 
    and then we compute it with Babylonian method. Both the approximation and Babylonian method are written as arithmetic equations without any if/else constructs.
\begin{algorithm}[tbp]
	\begin{algorithmic}[1]
		\caption{Normalized Secure Sum ($M$ party protocol)}
		\label{alg:nssum}
		\floatname{algorithm}{Procedure}
		\Require $\mbf x_m:$ party $m$'s input, a vector of cardinality $d$.
		\Ensure $\boldmath{\hat s} = \frac{\mbf s}{||\mbf s||_2}$, where $\mbf s=\sum_{m=1}^M \mbf x_m$
		\For{$m\in\{1,2,\dots, M\}$:}
		\For{$i \in {1,2, \dots, d}$:}
		\State $s[i] = s[i] + x_{m}[i]$
		\EndFor
		\EndFor
		\For{$i \in {1,2, \dots, d}$:}
		\State $s_{squared} = (s[i])^2$
		\EndFor
		
		\State $n = \sqrt{s_{squared}}$
		\For{$i \in {1,2, \dots, d}$:}
		\State ${\hat s}[i] = \frac{s[i]}{n}$
		\EndFor
	\end{algorithmic}
\end{algorithm}


%
%
%
%

\para{Security.} The security of our implementation follows directly from the security of the underlying SPDZ compiler. Indeed, other than the output $\mathbf{\hat s}$ which is reconstructed in the end, no other intermediate state is reconstructed in our implementation; hence, since SPDZ leaks no information during the  computation ~\cite{cryptoeprint:2011:535} neither does our implementation. 

\para{Numerical Precision} Since \nssum works by converting floats to fixed precision we must determine how many bits to allocate to the integer part and how many bits to the fractional part.
In order to simplify the problem the parties can scale each of their matrices by an upper bound for $||X^TX||_2$.
A simple upper bound is \[
||X^TX||_2 \leq ||(X\m)^TX\m||_F\equiv s,
\]
which can be computed using \ssum. By scaling their matrices by $\frac{1}{s}$ the parties ensure that in the power iterations, i.e. step 6 of Algorithm~\ref{alg:pdsvd}, $||S\mbf x||_2\leq 1$ since $||\mbf x||_2=1$. Therefore $||S\mbf x||_1\leq 1$, and as a result we know that the integer part will always be 0. Then we have reduced the problem to studying how many bits to allocate to the fractional part.
If we can make assumptions about measurement errors for the entries in $X$, it is possible to formally derive how many bits are needed~\cite{duryea1987finite}.
However for our experiments we don't make such assumptions, and simply compare using 10, 20, and 31 bits for the fractional part.

\para{Efficiency of \nssum} In Table~\ref{fig:nssum_time} we report the total wall-time that it takes to compute the normalized secure sum for vectors of various $d$ among various amounts of parties $M$. This includes conversion to and from fixed-point representation as well as the online phase in the SPDZ framework. The cost of the offline phase is not included since it can be run without access to $X\m$. 
This means the objects pre-computed in the offline phase can be used for diverse MPC applications (although they cannot be re-used).

\begin{table}[thbp]
\begin{tabular}{lrrr}
$M$&\multicolumn{3}{c}{$d$}\\
 
 &     100   & 1000       &   10000     \\
\midrule
2 &    1.4 &   11.5 &  113.9 \\
3 &    2.1 &   17.4 &  168.1 \\
4 &    3.5 &   29.3 &  293.0 \\
5 &    4.6 &   40.3 &  403.9 \\
\end{tabular}
\caption{Execution time in seconds per one invocation of \nssum for various vector dimensionality $d$ and number of parties $M$.\label{fig:nssum_time}}
\end{table}

For $d=100$, typical for PCR, the cost per iteration is about 2 sec with $M=3$.
This means $\tau=300$ iterations of PCR with $k=7$ can be computed in roughly 1.2 hours using \pdsvd.
For LRA $d=10000$ may be more typical, in which case the same would take roughly 4 days.
For the latter case if such a run-time is too long, \pdsvd can be run with \ssum replacing \nssum (in step 6 of Algorithm~\ref{alg:pdsvd}) to make communication overhead smaller; then \pdsvd could compute the same LRA in less than 30 minutes.
To inform this trade-off, we examine the effect of revealing $\Sigma$, and hence the privacy benefit of \nssum over \ssum, in the next section.